\documentclass{article} 
\usepackage{paper_conference,times}

\usepackage{amsmath,amsfonts,bm}

\def\eqref#1{equation~\ref{#1}}

\def\1{\bm{1}}

\DeclareMathAlphabet{\mathsfit}{\encodingdefault}{\sfdefault}{m}{sl}
\SetMathAlphabet{\mathsfit}{bold}{\encodingdefault}{\sfdefault}{bx}{n}

\usepackage{listings} 
\usepackage{geometry}
\usepackage{algorithm}
\usepackage{enumitem}
\usepackage{xspace}
\usepackage{algpseudocode} 
\usepackage{colortbl}
\usepackage{authblk}
\usepackage{hyperref}
\usepackage{url}
\usepackage[pdftex]{graphicx}
\usepackage{arydshln}
\usepackage{booktabs}
\usepackage{multirow}
\usepackage{wrapfig}
\usepackage{ulem}
\definecolor{cerisepink}{rgb}{0.93, 0.23, 0.51}
\newcommand{\ours}{{\scshape WALL-E}\xspace}

\title{\centering
WALL-E: \underline{W}orld \underline{A}lignment by Ru\underline{l}e \underline{Le}arning \\Improves World Model-based LLM Agents}

\author[1]{Siyu Zhou} 
\author[2]{Tianyi Zhou}  
\author[3]{Yijun Yang}
\author[1]{Guodong Long}
\author[3]{Deheng Ye}
\author[1]{Jing Jiang}
\author[1]{Chengqi Zhang}

\affil[1]{Australian AI Institute, Faculty of Engineering and IT, University of Technology Sydney}
\affil[2]{Department of Computer Science, University of Maryland, College Park}
\affil[3]{Tencent}

\affil[ ]{\texttt{Siyu.Zhou-2@student.uts.edu.au, zhou@umiacs.umd.edu}}

\affil[ ]{\textcolor{cerisepink}{Project: \url{https://github.com/elated-sawyer/WALL-E}}}

\begin{document}

\maketitle

\vspace{-0.9cm} 
\noindent

\begin{figure}[h]
  \vspace{0.0cm}
  \centering
  \includegraphics[width=1.0 \textwidth]{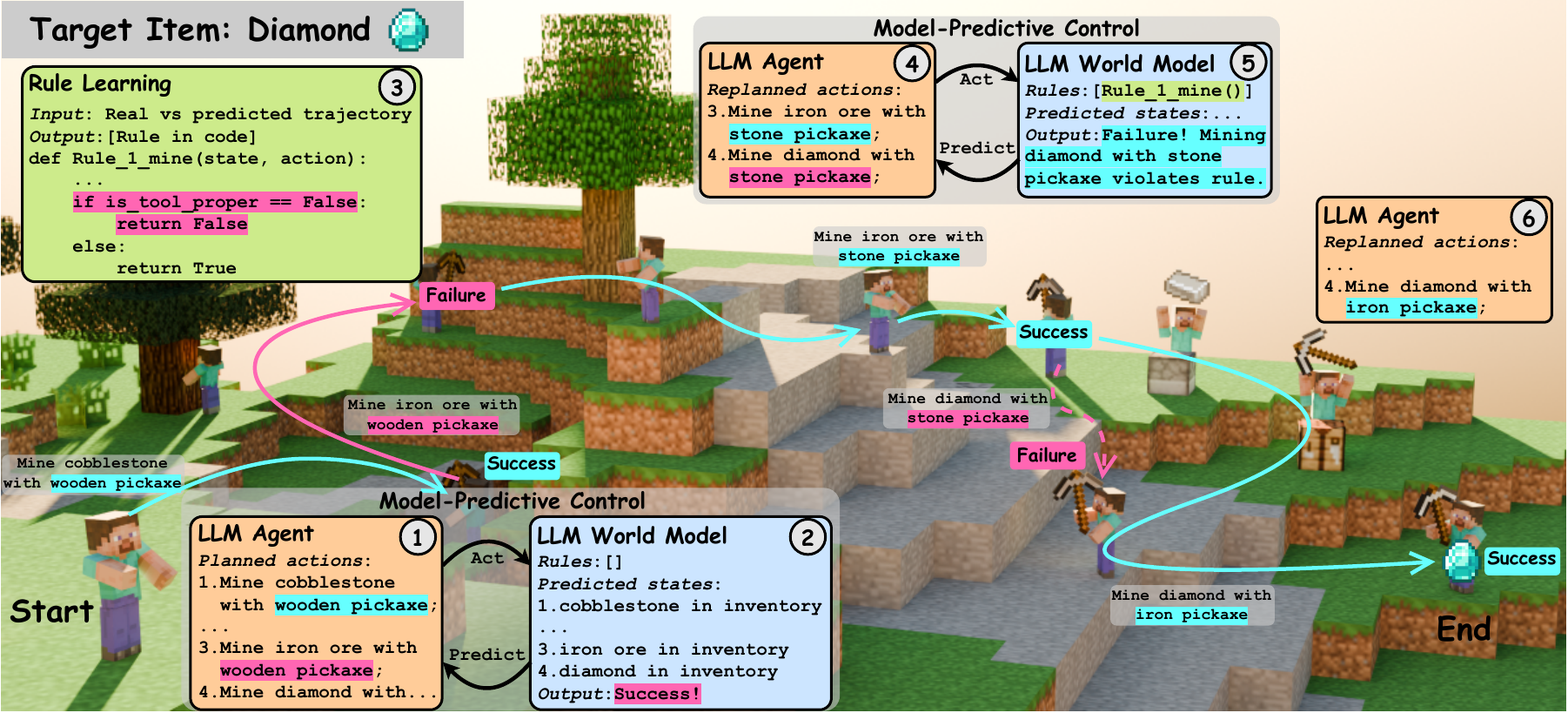} 
  \caption{\textbf{Illustration of WALL-E mining a diamond in Minecraft.}
  Step 1-2: the agent makes a plan via MPC with the initial unaligned world model, resulting in a failed action for mining iron ore. Step 3: by comparing real trajectories with the world model predictions, \ours learns a critical rule that if the tool is not proper to the material being mined, the action will fail. Step 4-5: the learned rule helps the world model make accurate predictions for transitions that were predicted mistakenly in MPC. Step 6: the agent accordingly modifies its plan and replaces \textit{stone pickaxe} with an \textit{iron pickaxe} toward completing the task.\looseness-1}
  \label{fig:teaser}
  \vspace{-0.2cm}
\end{figure}

\begin{abstract}
\textit{Can large language models (LLMs) directly serve as powerful world models for model-based agents?} While the gaps between the prior knowledge of LLMs and the specified environment's dynamics do exist, our study reveals that the gaps can be bridged by aligning an LLM with its deployed environment and such ``\textit{world alignment}'' 
can be efficiently achieved by rule learning on LLMs. Given the rich prior knowledge of LLMs, only a few additional rules suffice to align LLM predictions with the specified environment dynamics. 
To this end, we propose a neurosymbolic approach to learn these rules gradient-free through LLMs, by inducing, updating, and pruning rules based on comparisons of agent-explored trajectories and world model predictions. The resulting world model is composed of the LLM and the learned rules. Our embodied LLM agent ``\ours'' is built upon model-predictive control (MPC). By optimizing look-ahead actions based on the precise world model, MPC significantly improves exploration and learning efficiency. Compared to existing LLM agents, \ours's reasoning only requires a few principal rules rather than verbose buffered trajectories being included in the LLM input. On open-world challenges in Minecraft and ALFWorld, \ours~achieves higher success rates than existing methods, with lower costs on replanning time and the number of tokens used for reasoning. In Minecraft, WALL-E exceeds baselines by \textbf{15}-\textbf{30\%} in success rate while costing \textbf{8}–\textbf{20} fewer replanning rounds and only \textbf{60}–\textbf{80\%} of tokens. In ALFWorld, its success rate surges to a new record high of \textbf{95\%} only after \textbf{6} iterations. \looseness-1
\end{abstract}

\section{Introduction}
\vspace{-0.4cm}

\begin{wrapfigure}[26]{r}{0.4\textwidth}
\vspace{-0.5cm}
\centering
\includegraphics[width=0.4\textwidth]{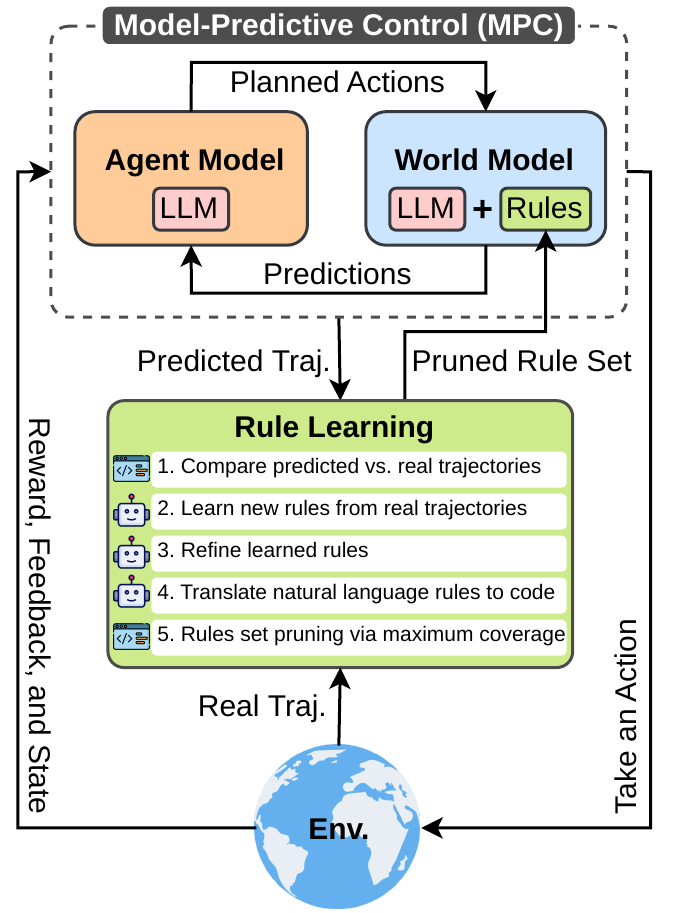}
\vspace{-0.75cm}
\caption{\textbf{Overview of \ours}. The agent's action per step is controlled by MPC, where the agent model plans actions in a look-ahead window based on the LLM+rules world model's predictions. \looseness-1
}
\label{fig:overview framework} 
\vspace{-0.8cm}
\end{wrapfigure}

While large language models (LLMs) have been successfully applied to complex reasoning, generation, and planning tasks, they are not sufficiently reliable to be deployed as an agent in specific open-world environments, e.g., games, VR/AR systems, medical care, education, autonomous driving, etc~\citep{gpt-4,nips/cot, liu2024aligning}. A primary reason for the failures is the gap between the commonsense reasoning with prior knowledge of pretrained LLMs and the specified, hard-coded environment's dynamics, which leads to incorrect predictions of the future states, hallucinations, or violation of basic laws in LLM agents' decision-making process~\citep{corr/embodiedgpt,yang2024embodied,cvpr/EmbodiedQA,wu2024autohallusion}. 
Although the alignment of LLMs with human preferences has been widely studied as a major objective of LLM post-training, ``world alignment'' with an environment's dynamics has not been adequately investigated in building LLM agents~\citep{hao2023reasoning,rafailov2024direct,ge2024worldgpt}. Moreover, many existing LLM agents are model-free and their actions are directly executed in real environments without being verified or optimized in advance within a world model or simulator~\citep{corr/embodiedgpt,yao2023react,shinn2024reflexion,corr/rt-2,wu2023autogen,micheli2021language,rss/rt-1}. This leads to safety risks and suboptimality of generated trajectories. \looseness-1

In this paper, we show that aligning an LLM with environment dynamics is both necessary and crucial to make it a promising world model, which enables us to build more powerful embodied agents. In particular, we introduce a neurosymbolic world model that composites a pretrained LLM with a set of newly learned rules from the interaction trajectories with the environment. This specific form of world model combines the strengths of both in modeling the environment dynamics, i.e., (1) the rich prior knowledge, probabilistic, and deductive reasoning capability of LLMs~\citep{hu2023language}; and (2) the hard constraints and rigorous guarantees enforced by rules~\citep{li2024ruler}. While creating a rule-only world model for a complex environment is challenging due to the massive amount of rules and uncertainty~\citep{xiao2021rule}, in our method, only a few complementary rules suffice to align a pretrained LLM to specific environment dynamics. This is achieved by simply including these rules in the LLM's prompt without tedious training or inference. 
In contrast, existing LLM agents usually require expensive finetuning of LLMs via RL/imitation learning on trajectory data, or memory-heavy inference with a long input context of buffered trajectories~\citep{corr/embodiedgpt,gao2023pg-vlm,yang2024embodied,shinn2024reflexion}. 

To this end, we propose ``\underline{W}orld \underline{A}lignment by ru\underline{L}e \underline{LE}arning (\ours)'', which builds the neurosymbolic world model by learning complementary rules with LLMs' inductive reasoning and code generation capability. Specifically, in each iteration, \ours interacts with the environment to collect a real trajectory and compare it with the world model predictions. The comparison results are then analyzed by an LLM, which extracts new rules or modifies existing ones to improve the consistency between the predicted and real trajectories. To keep the rule set minimal necessarily, at the end of each iteration, we prune the rules by solving a maximum coverage problem, which aims to select a subset of rules with the maximal coverage of the transitions failed being predicted by the LLM in the world model (without applying any rules). Hence, the selected rules are complementary to the LLM predictions. The above rule learning procedure repeats for multiple iterations until the LLM+rules performs as an accurate world model.  

The precise world model achieved by \ours enables us to create better model-based LLM agents for challenging open-world tasks.  
However, model-based reinforcement learning (RL) of LLM agents in complex environments is still hindered by the expensive exploration and finetuning of LLMs. 
In this paper, we revisit the classical idea of model-predictive control (MPC)~\citep{qin2003survey,hafner2019dream,hafner2020mastering,hafner2023mastering}, compared to RL, which does not require training a policy network but needs to optimize actions for a look-ahead time window in every step. 
To reduce the optimization cost per step, we instead apply the LLM agent as an optimizer searching for the optimal look-ahead actions by interacting with the \ours's world model. With an aligned world model and an efficient LLM-based optimizer, MPC leads to a more promising and efficient framework of LLM agents in open-world environments. 

We evaluate \ours on challenging tasks in open-world environments such as Minecraft and ALFWorld where the agents can explore freely and target complicated tasks. Our main contributions are threefold. \looseness-1
\begin{itemize}[leftmargin=1em]
    \item We investigate the underexplored ``world alignment'' challenge for LLM agents. 
    \item We propose a novel class of neurosymbolic world models based on rule learning on LLMs. 
    \item We develop LLM agents based on model-predictive control (MPC) with the neurosymbolic world model.  
\end{itemize}

\section{Related Work}

Recent studies have integrated LLMs with rule learning to improve reasoning and generalization capabilities across various tasks, including numerical reasoning, knowledge graph exploration, and adherence to predefined rules~\citep{yang2023enabling, Zhu2023Large, mu2023can, yang2023failures, luo2023chatrule}. However, prior work has not focused on aligning LLM-based world models with dynamic environments. Our research addresses this gap by applying rule learning to enhance model-based agent performance in such contexts.
Several works have also used LLMs to construct world models for task planning by translating natural language into representations or combining LLMs with task-specific modules~\citep{wong2023learning, guan2023leveraging, tang2024worldcoder}. Unlike these approaches, we directly employ LLMs as world models, leveraging their inherent knowledge for greater flexibility and efficiency.
While some works use LLMs as world models, typically relying on fine-tuning or human defined prompts for alignment with environment dynamics~\citep{xiang2024language, xie2024making, zhao2024large, hao2023reasoning, liu2023reason}. Our method advances this by automatically learning rules through exploration, reducing human intervention and improving performance. 
For a more comprehensive discussion of related work, please refer to Appendix \ref{sec:Detailed Related Work}.

\section{Method}

\subsection{Model-Predictive Control (MPC) of World Model-based LLM Agents}

We consider a scenario where a LLM, denoted as $f$, is deployed in a dynamic environment for agent interaction over discrete time steps. At each time step $t$, the agent observes the current state $s_t$, selects an action $a_t$, and transitions to the next state $s_{t+1}$. This transition is represented as $\delta_t = (s_t, a_t, s_{t+1})$. A trajectory $\tau = (\delta_0, \delta_1, \ldots, \delta_{T-1})$ comprises a sequence of such transitions, capturing the agent's behavior from the initial to the terminal state within an episode.

The LLM-based world model $f_{\text{wm}}$ predicts the subsequent state $\hat{s}_{t+1}$ based on the current state and action:
\begin{equation}
    \hat{s}_{t+1} = f_{\text{wm}}(s_t, a_t),
\end{equation}
Model Predictive Control (MPC) is a widely recognized framework for model-based control. In this context, we integrate MPC with the LLM-based world model $f_{\text{wm}}$ to enhance agent planning and decision-making, the whole framework is illustrated in Figure \ref{fig:overview framework}. 
The objective is to determine an optimal sequence of actions $a_{t:t+H}$ over a finite horizon $H$that maximizes the expected cumulative reward. At each time step $t$, the optimization problem is formulated as:
\begin{equation}
a_{t:t+H}^* = \arg \max_{a_{t:t+H}} \mathbb{E} \left[ \sum_{i=0}^{H} \gamma^{i} \mathcal{F}(\hat{s}_{t+i+1}) \right],
\end{equation}
where $\gamma$ is the discount factor, and $\mathcal{F}(\hat{s}_{t+i+1})$ denotes the reward function. 

However, if the LLM-based world model is misaligned with the actual environment dynamics, the predicted state $\hat{s}_{t+1}$ may not match the true state $s_{t+1}$. This misalignment leads to incorrect reward evaluations, resulting in inaccurate cumulative reward estimates. Consequently, the derived action sequence $a_{t:t+H}^*$ may be suboptimal or erroneous, leading to ineffective control decisions by the agent.
Therefore, addressing the misalignment between the LLM world model and the environment's true dynamics is crucial for ensuring optimal performance within the MPC framework.

\subsection{World Alignment by Rule Learning (\ours)}

\begin{figure}[t!]
\begin{center}
\includegraphics[width=0.9 \linewidth]{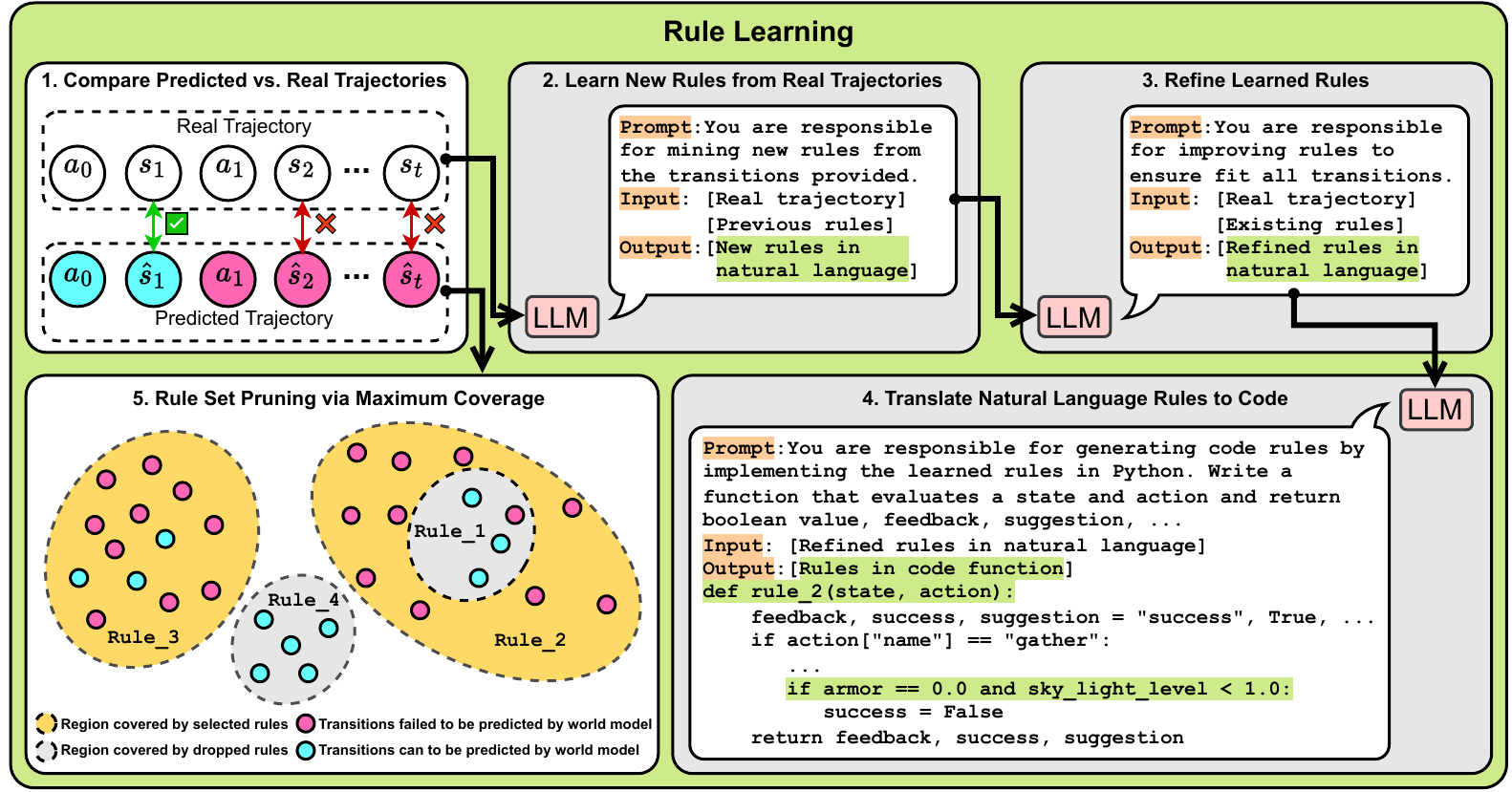} 
\end{center}
\vspace{-1em}
\caption{\textbf{Rule Learning details}.
The rule learning module iteratively refines the rules by comparing the world model predicted trajectories with the agent's actual trajectories in the environment. 
The rule learning takes five steps: (1) comparing predicted and actual trajectories; (2) learning new rules from real trajectories; (3) refining learned rules; (4) translating natural language rules to code; and (5) rule set pruning via solving a maximum coverage problem. (2)-(4) are handled by LLMs, while (1) and (5) are executed by programs. }
\label{fig:Rule Learning Framework} 
\end{figure}

In complex environments, direct state prediction is challenging due to complexity and randomness. To address this, our world model uses a two-stage approach: first, it assesses $\text{action\_result}$ (e.g., success or failure), then generates the subsequent $\text{state\_info}$ (provides state details) based on the action success:
\begin{equation}
\hat{s}_{t+1} = \left( \text{action\_result}_{t+1}, \text{state\_info}_{t+1} \right) = f_{\text{wm}}(s_t, a_t),
\end{equation}

To address potential misalignment between the $f_{\text{wm}}$ and the real environment, we introduce a rule learning framework, illustrated in Figure \ref{fig:Rule Learning Framework} and
detailed in the following sections.
The learned rules align the $f_{\text{wm}}$ with the environment, enhancing state prediction accuracy and improving agent performance within the MPC framework.

\paragraph{Comparing Predicted and Real Trajectories.}
\label{sec:Comparing Predicted and Real Trajectories}
To find misalignments between the LLM world model and the real environment, we compare action outcomes in predicted and actual next state, focusing on the binary $\text{action\_result}$ rather than detailed $\text{state\_info}$. This focus provides a reliable basis for identifying discrepancies.
Let the predicted trajectories be $\tau^{\text{predicted}} = \{ \delta = (s_t, a_t, \hat{s}_{t+1}) \}_{t=0}^{T}$. 
Then, we may divide $\tau^{\text{predicted}}$ into correct and incorrect transition set, and correct the wrong $\hat{s}_{t+1}$ (see Step 1 of rule learning in Figure \ref{fig:Rule Learning Framework}):
\begin{equation}
    \begin{aligned}
        \mathcal{D}^{\text{correct}} &= \left\{ \delta^{\text{correct}}_t = (s_t, a_t, \hat{s}_{t+1}) \; \middle|\; \hat{s}_{t+1} = s_{t+1} \right\}, \\
        \mathcal{D}^{\text{incorrect}} &= \left\{ \delta^{\text{incorrect}}_t = (s_t, a_t, s_{t+1}) \; \middle|\; \hat{s}_{t+1} \neq s_{t+1} \right\},
    \end{aligned}
\end{equation}
where $s_{t+1}$ is the true state given by environment.
Then $\tau^{\text{predicted}} = \mathcal{D}^{\text{correct}} \cup \mathcal{D}^{\text{incorrect}}$.
By analyzing $\mathcal{D}^{\text{incorrect}}$, we pinpoint where the model's predictions diverge from reality, highlighting areas needing correction through additional rules.

\paragraph{Learning New Rules from Real Trajectories.}
Before address these misalignments, we prompt the LLM $f_{\text{gen}}$ to generate new natural language rules from real trajectories $\tau^{\text{real}}$ (see Appendix \ref{sec:Prompt for Learn New Rules from Real Trajectories} for detailed prompt).
The LLM is given the task setup and state-action structures to infer new natural language rules \( R_{\text{new}}^{\text{NL}} \) that explain the observed dynamics, ensuring they are distinct from previous rules \( R_{\text{previous}}^{\text{NL}} \):
\begin{equation}
    R_{\text{new}}^{\text{NL}} = f_{\text{gen}}(\tau^{\text{real}}, R_{\text{previous}}^{\text{NL}}),
\end{equation}

\paragraph{Refining Learned Rules.}

Then, we prompt the LLM to update existing rules based on the real trajectories $\tau^{\text{real}}$ (see Appendix \ref{sec:Refine Learned Rules} for detailed prompt). Early-stage rules could be inaccurate due to data drift caused by the limited data, so the LLM identifies conflicting rules and modifies or discards them as needed. The set of all existing rules up to the current point is $R_{\text{existing}}^{\text{NL}} = R_{\text{previous}}^{\text{NL}} \cup R_{\text{new}}^{\text{NL}} $, where the LLM $f_{\text{refine}}$ refines these rules with the real trajectories:
\begin{equation}
    R^{\text{NL}} = f_{\text{refine}}(\tau^{\text{real}}, R_{\text{existing}}^{\text{NL}}).
\end{equation}

\paragraph{Translating Natural Language Rules to Code.}
\label{sec:Translating Natural Language Rules to Code}

The next step is translating the refined natural language rules $R^{\text{NL}}$ into executable code. 
We prompt the LLM $f_{\text{code\_gen}}$ to produce the code-based rule set $R^{\text{code}}$   (see Appendix \ref{sec:Translate Natural Language Rules to Code} for detailed prompt):
\begin{equation}
    R^{\text{code}} = f_{\text{code\_gen}}(R^{\text{NL}}),
\end{equation}

\paragraph{Rule Set Pruning via Maximum Coverage.}
\label{sec:Rule Set Pruning via Maximum Coverage}

In the final step, to address the inherent uncertainty and variability in the LLM-driven rule-learning process, we programmatically verify and refine the rule set to reduce dependence on the LLM. The code-based rules $R^{\text{code}}$ are executed and validated against the labeled predicted transitions $\tau^{\text{predicted}}$. Any rule that fails to predict a transition correctly is discarded, ensuring that only accurate and effective rules are retained.

We further optimize the rule set by selecting rules that maximize coverage of the incorrectly predicted transitions $\delta^{\text{incorrect}}_t$, where the LLM world model's failed. This approach focuses our efforts on correcting the most significant misalignments between the LLM and the environment.
We formulate this optimization as a maximum set cover problem. \( \mathcal{D}^{\text{incorrect}} = \{ \delta^{\text{incorrect}}_1, \delta^{\text{incorrect}}_2, \ldots, \delta^{\text{incorrect}}_n \} \) is the set of incorrectly predicted transitions, and \( R^{\text{code}} = \{ R^{\text{code}}_1, R^{\text{code}}_2, \ldots, R^{\text{code}}_m \} \) is the set of code-based rules. Our goal is to select a minimal subset of rules that maximizes coverage of \( \mathcal{D}^{\text{incorrect}} \):
\begin{equation}
\label{eqn:maximum set cover problem}
    \max_{\bm{x} \in \{0,1\}^{m},\; \bm{y} \in \{0,1\}^{n}} \left\{ \sum_{j=1}^{n} y_j - \lambda \sum_{i=1}^{m} x_i  \ \bigg\vert\ y_j \leq \sum_{i=1}^{m} x_i a_{ij},\ \forall j = 1, \ldots, n \right\},
\end{equation}
where \( x_i \) indicates whether rule \( R^{\text{code}}_i \) is selected (\( x_i = 1 \)) or not (\( x_i = 0 \)), \( y_j \) indicates whether transition \( \delta^{\text{incorrect}}_j \) is covered (\( y_j = 1 \)) or not (\( y_j = 0 \)), and \( a_{ij} = 1 \) if transition \( \delta^{\text{incorrect}}_j \) is covered by rule \( R^{\text{code}}_i \), \( a_{ij} = 0 \) otherwise. 
The constraint ensures that a transition $\delta^{\text{incorrect}}_j$ is considered covered if it is included in at least one selected rule.
The parameter \( \lambda > 0 \) balances minimizing the number of rules and maximizing transition coverage; we set \( \lambda \) to be very small to prioritize coverage maximum. We solve this optimization problem using a greedy algorithm (see Appendix~\ref{sec:Greedy Algorithm}).

Through this process, we eliminate \textbf{rules covering only correct transitions}, as they do not address misalignments, and \textbf{redundant rules} fully covered by more comprehensive ones (see Step 5 of rule learning in Figure \ref{fig:Rule Learning Framework}). 
This pruning process results in a pruned rule set that is both efficient and effective in correcting the LLM's misalignments. Additionally, any code-based rules removed from $R^{\text{code}}$ are also excluded from the set of natural language rules $R^{\text{NL}}$.

\subsection{Inference on LLM Agents with Learned Rules}
\label{sec:Inference on LLM Agents with Learned Rules}

After completing the rule learning process, we obtain rules in two distinct forms: natural language rules $R^{\text{NL}}$ and code-based rules $R^{\text{code}}$. Both types of rules enhance the LLM world model’s ability to predict the next state $\hat{s}_{t+1}$ within the planning framework:
For \textbf{natural language rules}, these can be embedded directly into the LLM's input prompt to guide the model’s predictions, e.g., $\hat{s}_{t+1} = f_{\text{wm}}(s_{t}, a_{t}, R^{\text{NL}})$. 
For \textbf{code-based rules}, these are applied programmatically after the LLM generates its initial prediction, e.g., $\hat{s}_{t+1} = \text{ApplyRules}(f_{\text{wm}}(s_{t}, a_{t}), R^{\text{code}})$. Here, the function $\text{ApplyRules}$ serves as a verification layer, overriding the LLM's prediction if an active rule contradicts the generated outcome. For further details on rule activation, refer to Appendix \ref{sec:verification logic}.

By integrating learned rules, the aligned LLM world model enhances the agent's planning process significantly. This alignment allows the agent to more effectively obtain optimal action sequences $a_{t:t+H}$ through two key improvements:
First, the alignment leads to more \textbf{accurate reward evaluations $\mathcal{F}(\hat{s}_{t+1})$}, increasing the likelihood of selecting optimal action sequences $a_{t:t+H}$ within the MPC framework. 
Second, the aligned world model, equipped with learned rules, provides \textbf{high-quality feedback} that helps the agent refine $a_{t:t+H}$ effectively. Along with predicting action results and state information, it offers auxiliary information when an action is predicted to fail, including:
\begin{itemize}
    \item Feedback: A textual explanation of the failure based on violated rules.
    \item Suggestion: Recommendations for corrective actions or improvements based on the current state, action taken, and violated rules.
\end{itemize}
This information is crucial when an action fails, guiding the agent in revising its strategy by exploring alternatives or adjusting its approach(see Appendix \ref{sec:Code-based Rules} for examples).

In conclusion, integrating learned rules improves the LLM world model's prediction accuracy and provides actionable feedback, enabling more efficient and adaptive planning.

\vspace{-0.2cm}
\section{Experiments}
\vspace{-0.2cm}

\begin{wrapfigure}[16]{r}{0.4\textwidth}
\vspace{-0.4cm}
\centering
\includegraphics[width=0.4\textwidth]{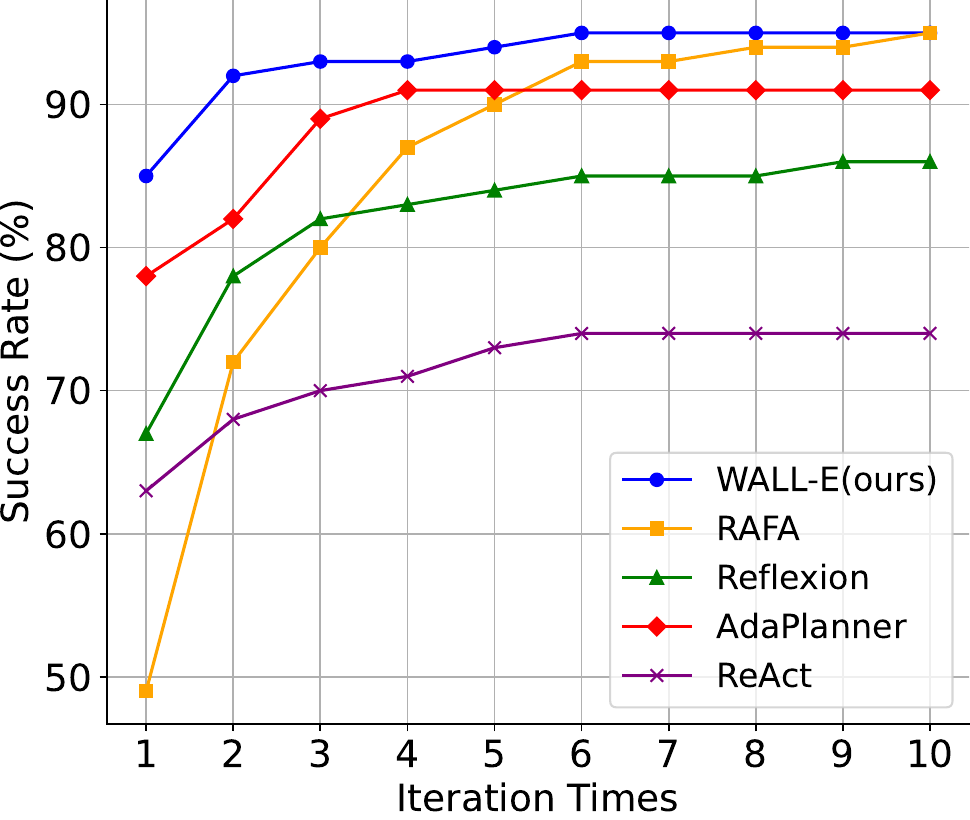}
\vspace{-0.5cm}
\caption{Comparison of \ours and baselines on 134 testing tasks from the ALFWorld benchmark.}
\label{fig:Alfworld_Success_Rate_Across_Episodes}
\vspace{-0.8cm}
\end{wrapfigure}

We evaluate the environment modeling and task-solving capabilities of WALL-E on open-world environments using the Minecraft \citep{fan2022minedojo} and ALFWorld \citep{shridhar2020alfworld} benchmarks. Compared to state-of-the-art (SOTA) LLM/VLM agents, WALL-E achieves higher success rates with lower costs in terms of replanning time and token usage for reasoning. Notably, in Minecraft, WALL-E surpasses baselines by 15–30\% in success rate while costing 8–20 fewer replanning rounds and only 60–80\% of tokens. In ALFWorld, it achieves a record of 95\% success rate only after 6 iterations, significantly exceeding other planning-based methods such as RAFA~\citep{liu2023reason}. 
Moreover, integrated with our proposed rule learning method, WALL-E achieves a 15\% higher success rate than methods relying on a long input context of buffered trajectories. These highlights demonstrate WALL-E's superior efficiency and effectiveness in complex and open-world environments.

\subsection{Experimental Setup}

\paragraph{Benchmarks.}\textbf{Minecraft} is a popular open-world environment.
We employ the standard evaluation pipeline provided by MineDojo's TechTree tasks \citep{fan2022minedojo}. These tasks can be categorized into six levels of increasing difficulty: \textit{Wood}, \textit{Stone}, \textit{Iron}, \textit{Gold}, \textit{Diamond}, and \textit{Redstone} (see Appendix \ref{sec:Minecraft Experiment Details} for details).
\textbf{ALFWorld} is a virtual environment designed as a text-based simulation where agents perform tasks by interacting with a simulated household environment~\citep{shridhar2020alfworld}. This benchmark includes six distinct task types, each requiring the agent to accomplish a high-level objective, such as placing a cooled lettuce on a countertop  (see Appendix \ref{sec:ALFWorld Experiment Details} for details).\looseness-1

\paragraph{Metrics.}(1) \textbf{Success rate} (higher is better): the percentage of tasks the agent completes successfully. (2) \textbf{Replanning rounds} (lower is better): the number of times the agent revisits the same task to revise its plan for recovering from the failed task planning. (3) \textbf{Token cost} (lower is better): the number of tokens consumed by LLM agent/world models during task completion.
For Minecraft, we select four tasks from each level to serve as the testing set and the remaining tasks to construct the training set. 
All these three metrics are employed in our experiment.
The task will be marked incomplete if the agent either dies in the environment (such as being killed by hostile mobs or falling into lava) or reaches one of the following maximal budgets: 10-minute time limit and maximum replanning rounds. In these cases, the replanning rounds and token cost will be set to the maximal value. 
For ALFWorld, we train WALL-E on the designated training set and evaluate its performance on a set of 134 predefined testing tasks. The averaged success rate over several trials is used as the evaluation metric to measure the performance of all baselines.

\subsection{Main Results}

\begin{table}[t]
\caption{Comparison of WALL-E and baselines on Minecraft tasks for success rate (\%) and replanning rounds. $*$-reported in previous work.
VLMs = vision-language models, LLMs = large language models. The best score for each task is highlighted in \textbf{bold}.
\textbf{WALL-E substantially exceeds other SOTA LLM/VLM agents and is the only method that performs better than human players in the Minedojo benchmark}.
}
\vspace{-0.5em}
\label{tab:success rate and replanning rounds}
\begin{center}
\renewcommand\arraystretch{1.1}
\resizebox{\linewidth}{!}{

    \begin{tabular}{llccccccc}
    \toprule
    \multicolumn{2}{l}{\multirow{2}{*}{\textbf{Method}}}     & \multicolumn{7}{c}{\textbf{Success Rate (\%) $\uparrow$ (Replanning Rounds $\downarrow$)}} \\ \cmidrule(lr){3-9} 
    \multicolumn{2}{l}{}                                                    & Avg.           & Wooden          & Stone           & Iron           & Golden          & Diamond          & Redstone \\ \midrule
    \multirow{3}{*}{\rotatebox{90}{\textbf{VLMs}}} & GPT-4V*~\citep{li2024optimus}                 & 10(-)          & 41(-)           & 21(-)           & 0(-)           & 0(-)            & 0(-)             & 0(-)     \\
                                                   & Jarvis-1*~\citep{wang2023jarvis}  & 42(-)          & 94(-)           & 89(-)           & 36(-)          & 7(-)            & 9(-)             & 16(-)    \\
                                                   & Optimus-1*~\citep{li2024optimus} & \textbf{47}(-) & \textbf{99}(-)  & \textbf{92}(-)  & \textbf{47}(-) & \textbf{9}(-)   & \textbf{12}(-)   & \textbf{25}(-)    \\ \midrule
    \multirow{5}{*}{\rotatebox{90}{\textbf{LLMs}}} & GPT-3.5*~\citep{li2024optimus}                & 10(-)     & 40(-)  & 20(-)  & 0(-)  & 0(-)   & 0(-)    & 0(-)     \\
                                                   & DEPS~\citep{wang2023describe}                   & 37(35.36) & 83(10.67) & 41(33.26)          & 33(35.27) & 22(45.29) & 24(42.46) & 17(45.22)\\
                                                   & GITM~\citep{zhu2023ghost} & 54(25.49) & 96(3.42)  & \textbf{92}(6.01)  & 57(23.93)      & 29(37.17)     & 30(39.80)      & 22(42.63)\\
                                                   & WALL-E w/o WM          & 61(23.13)        & 94(5.04)         & 89(9.58)     & \textbf{67}(\textbf{18.56})  & 33(39.67)      & 41(32.73)          & 43(33.21)\\
                                                   & \textbf{WALL-E (ours)} & \textbf{69}(\textbf{15.77}) & \textbf{98}(\textbf{1.64})   & 91(\textbf{4.58})   & 63(19.38)     & \textbf{69}(\textbf{15.61})  & \textbf{46}(\textbf{27.08})  & \textbf{48}(\textbf{26.33})  \\ \midrule
    \multicolumn{2}{l}{\textbf{Human Performance}*~\citep{li2024optimus}}          & 59(-) & 100(-) & 100(-) & 86(-) & 17(-)  & 17(-)   & 33(-)    \\ \bottomrule
    \end{tabular}
    }
\end{center}
\end{table}

\begin{table}[t]
\caption{Comparison of WALL-E and baselines on Minecraft tasks for average token usage and API costs  (in USD). The number of tokens is calculated as the sum of prompt tokens and generation tokens. The average API cost is derived by separately calculating the costs of prompt and generation tokens and then summing both. The lowest cost for each task is highlighted in \textbf{bold}.
}
\vspace{-0.5em}
\label{tab:token-cost-table}
\begin{center}
\renewcommand\arraystretch{1.1}
\resizebox{\textwidth}{!}{
    \begin{tabular}{lcccccccc}
    \toprule
    \multirow{2}{*}{\textbf{Method}} & \multicolumn{8}{c}{\textbf{Inference Tokens $\downarrow$ (Cost in USD $\downarrow$)}}          \\ \cmidrule(lr){2-9} 
                            & Avg. & Wooden        & Stone        & Iron         & Golden       & Diamond      & Redstone       \\ \midrule
    DEPS                    &93560.95(0.65) &28223.33(0.20) & 87999.46(0.61) & 93313.38(0.65) & 119827.88(0.84) & 112346.49(0.79) &119655.16(0.84)            \\
    GITM                &74638.54(0.51) &\textbf{10027.71}(\textbf{0.07}) & \textbf{17566.79}(\textbf{0.12}) & 70071.99(0.48) & 108816.53(0.74) & 116526.40(0.80)   & 124821.83(0.85) \\
    WALL-E w/o WM           &72390.16(0.52) &15759.72(0.11) & 29976.28(0.21) & 58074.70(0.41) & 124147.71(0.89) & 102447.94(0.73)   & 103934.58(0.74)  \\
    \textbf{WALL-E (ours)}           &\textbf{60348.71}(\textbf{0.41}) &23179.52(0.15)  & 36595.33(0.24)&\textbf{57106.20}(\textbf{0.39})&\textbf{84776.25(0.58)}&\textbf{59261.31(0.40)}&\textbf{101173.64(0.68)}\\ \bottomrule
    \end{tabular}
    }
\end{center}
\vspace{-0.5em}
\end{table}

We conduct a detailed comparison of WALL-E and existing baseline methods in Tables \ref{tab:success rate and replanning rounds}, \ref{tab:token-cost-table} and \ref{tab:alfworld-method-comparison-table}, to demonstrate its superior performance in terms of success rate, planning efficiency, and token cost consumed by LLMs across diverse tasks.

\textbf{WALL-E demonstrates superior planning and task-solving abilities.} 
Tables \ref{tab:success rate and replanning rounds} and \ref{tab:alfworld-method-comparison-table} show that our method achieves the highest success rates across different environments. Specifically, in the Minecraft environment, WALL-E outperforms other baselines by an impressive margin of 15–30\%.
Figure~\ref{fig:Alfworld_Success_Rate_Across_Episodes} shows that WALL-E achieves the highest success rate after only 6 iterations, significantly surpassing other SOTA planning-based baselines such as RAFA~\citep{hao2023reasoning} and AdaPlanner~\citep{sun2024adaplanner}. 

\textbf{Aligned world model leads to higher sample efficiency.} While the integration of the LLM world model leads to additional token costs compared to model-free methods, WALL-E demonstrates remarkably high sample efficiency, which is sufficient to offset the additional consumption caused by the world modeling. 
Specifically, our method requires 8–20 fewer replanning rounds than other baselines (see Table \ref{tab:success rate and replanning rounds}), resulting in overall token usage that is only 60–80\% of that observed in other methods (see Table \ref{tab:token-cost-table}).
It is worth noting that the advantage of WALL-E becomes more apparent in harder environments. In turn, model-free methods can only achieve comparatively high sample efficiency on those easy tasks such as \textit{Wood} and \textit{Stone}.

\textbf{WALL-E is a general and environment-agnostic method.}
Unlike methods tailored to specific environments, e.g., GITM~\citep{zhu2023ghost} for open-world exploration in Minecraft and BUTLER~\citep{micheli2021language} for long-horizon planning in ALFWorld, WALL-E can excel at both, underscoring its generalizability and effectiveness in enhancing agent's capabilities of exploration, planning, and reflection in general, complex scenarios.

\begin{table}[t]
\caption{Comparison of \ours and baselines on 134 testing tasks from the ALFWorld benchmark. $*$-reported in previous work. VLMs = vision-language models, LLMs = large language models. The success rate (\%) is the percentage of tasks completed successfully. The best score for each task is highlighted in \textbf{bold}. \looseness-1}
\vspace{-0.5em}
\label{tab:alfworld-method-comparison-table}
\begin{center}
\renewcommand\arraystretch{1.1}
\resizebox{0.8\linewidth}{!}{

    \begin{tabular}{llccccccc}
    \toprule
    \multicolumn{2}{l}{\multirow{2}{*}{\textbf{Method}}}     & \multicolumn{7}{c}{\textbf{Success Rate (\%) $\uparrow$}} \\ \cmidrule(lr){3-9} 
    \multicolumn{2}{l}{}                                                      & Avg.      & Pick      & Clean     & Heat      & Cool      & Examine   & Picktwo \\ \midrule
    \multirow{5}{*}{\rotatebox{90}{\textbf{VLMs}}} & MiniGPT-4*~\citep{zhu2023minigpt}     &16           &4           &0           &19           &17           &67           &6           \\
                                                   & BLIP-2*~\citep{li2023blip}        &4           &0           &  6         &  4         &  11         &  6         &  0         \\
                                                   & LLaMA-Adapter*~\citep{gao2023llama} & 13          & 17          & 10          & 27          & 22          & 0          &  0         \\
                                                   & InstructBLIP*~\citep{dai2023instructblip}  & 22          & 50          & 26          & 23          & 6          &  17         &  0         \\
                                                   & EMMA*~\citep{yang2024embodied}& \textbf{82}  & \textbf{71}  & \textbf{94}  & \textbf{85} & \textbf{83}  & \textbf{88} &  \textbf{67}         \\ \midrule
    \multirow{9}{*}{\rotatebox{90}{\textbf{LLMs}}} & BUTLER*~\citep{micheli2021language}        & 26          & 31          & 41          & 60          & 27          & 12          &  29         \\
                                                   & GPT-BUTLER*~\citep{micheli2021language}     & 69          & 62          & 81          & 85          & 78          & 50          &  47         \\
                                                   & DEPS~\citep{wang2023describe} & 76          & 93          & 50          &  80         & \textbf{100}          & \textbf{100}          & 0          \\
                                                   & AutoGen*~\citep{wu2023autogen} & 77          & 92          & 74          &  78         & 86          &  83         &  41         \\
                                                   & ReAct~\citep{yao2023react} & 74          & 79           & 54           & 96          & 85        & 83        & 51        \\
                                                   & AdaPlanner~\citep{sun2024adaplanner}& 91        & \textbf{100} & \textbf{100} & 89           & \textbf{100} & 97           & 47           \\
                                                   & Reflexion~\citep{shinn2024reflexion}& 86          & 92           & 94           & 70           & 81           & 90           & 88           \\
                                                   & RAFA~\citep{liu2023reason}& \textbf{95} & \textbf{100} & 97           & 91           & 95           & \textbf{100} & 82           \\
                                                   & \textbf{WALL-E (ours)}     &  \textbf{95} & \textbf{100} & 97           & \textbf{100} & 86           & 85           & \textbf{100} \\ \midrule
    \multicolumn{2}{l}{\textbf{Human Performance}*~\citep{cvpr/alfred}}          & 91 & - & - & - & -  & -   & -    \\ \bottomrule
    \end{tabular}
    }
\end{center}
\end{table}

\subsection{Effectiveness of Rule Learning}

\begin{figure}[t]
\begin{center}
\includegraphics[width=5.5in]{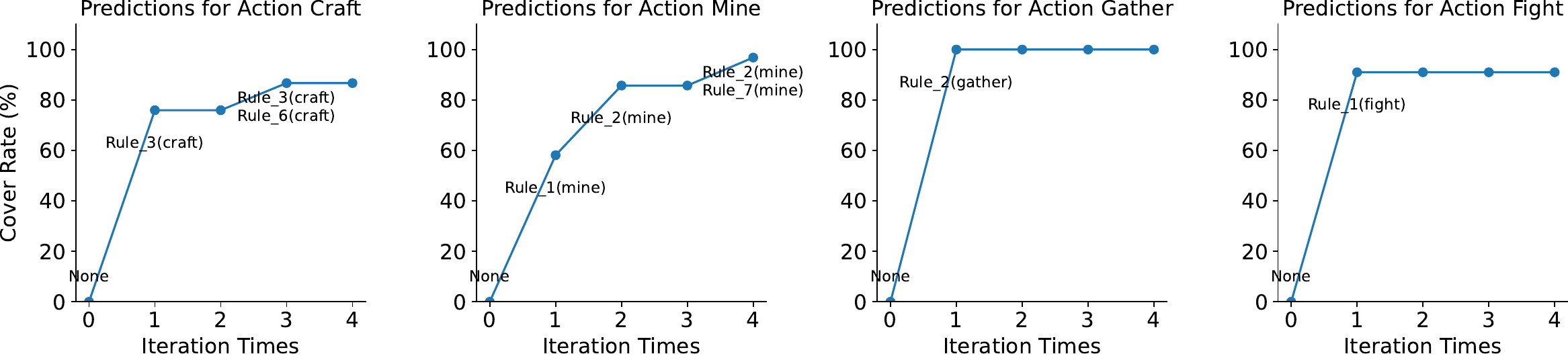} 
\end{center}
\vspace{-0.3cm}
\caption{Cover rate of LLM failed predictions across different actions over iteration times during training. The cover rate represents the probability that the LLM's failed predictions are addressed by the rules obtained during the rule learning process. The predictions and rules are categorized by action type: \textit{craft}, \textit{mine}, \textit{gather} and \textit{fight}. The learnt rules at each iteration are displayed in black under each node, labeled with their respective rule IDs. }
\label{fig:rulelearningprocess} 
\vspace{-0.3cm}
\end{figure}

\begin{figure}[t]
\begin{center}
\includegraphics[width=5in]{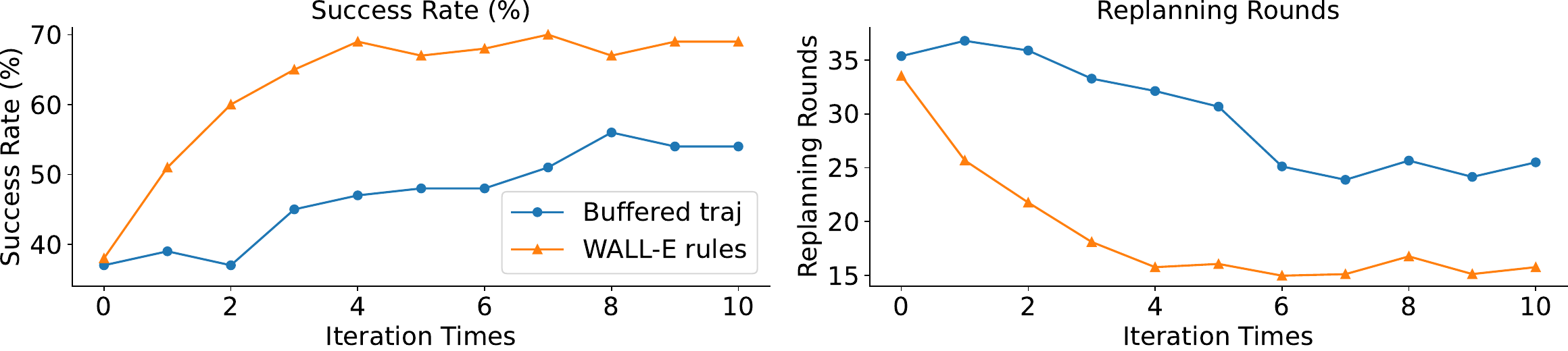} 
\end{center}
\vspace{-0.3cm}
\caption{Learning curve comparison between rule learning (e.g., WALL-E) and buffered trajectory (e.g., GITM) over 10 iterations on Minecraft tasks during training. The left plot shows the average success rate (\%) across all tasks, where a higher value indicates more tasks successfully completed. The right plot illustrates the average number of replanning rounds, with fewer rounds indicating higher efficiency in task completion. }
\label{fig:LearningCurve} 
\vspace{-0.2cm}
\end{figure}

In order to demonstrate the effectiveness of our proposed rule learning method, we conduct a comparative study against GITM~\citep{zhu2023ghost} - a method employing buffered trajectories as in-context examples to align LLM agents with the environment dynamics.
By jointly examining the rule learning process (Figure \ref{fig:rulelearningprocess}) and the agent's training progress (Figure \ref{fig:LearningCurve}), we observe an interesting phenomenon that WALL-E's success rate hits the upper bound after 4 iterations, while the rule learning process also finds a compact set of rules for the LLM world model and keeps this set fixed after 4 iterations, reflecting that WALL-E's improvement mainly benefits from the learning of new rules.

\textbf{Rule learning achieves efficient ``world alignment''.} To verify whether the learned rules enable a more accurate world model, we first collect a dataset of transitions that cannot be predicted by the LLM world model correctly and evaluate each rule on this dataset by calculating the cover rate - the probability that the LLM's failed predictions are addressed by the rules obtained during the rule learning process.
According to Figure \ref{fig:rulelearningprocess}, it is evident that the rules learned by our proposed framework consistently improve cover rates across different types of actions in the Minedojo benchmark. 
In specific, actions such as gather and fight reach 100\% and 91\% coverage after the first iteration, while craft and mine actions demonstrate improvements over multiple iterations, with final coverage rates of 87\% and 96\%, respectively.

\subsection{Ablation Study}

We conduct a comprehensive ablation study to evaluate the importance of various components in \ours. Specifically, we separately remove the learned rules and the world model and check their effects on WALL-E's final performance. \textbf{According to the results in Table \ref{tab:minecraft_ablation_study}, we give the following conclusions.}
(1) Regardless of whether the learned rules are applied within the agent or the world model, adding them significantly enhances the total performance. The success rate increases by 20\% to 30\% approximately. This observation underscores the crucial role that rules play in improving the effectiveness of \ours.
(2) When the learned rules are utilized within the world model, they contribute to nearly a 30\% improvement in success rate, whereas using rules within the agent result in about a 20\% improvement. This disparity may be primarily due to the fact that the learned rules are highly related to the state information (See Appendix \ref{sec:Learned Rules} for more details). 
(3) MPC using a world model without applying any rules cannot significantly improve \ours's performance in terms of the success rate and the number of replanning times. This finding suggests that the alignment between the world model and the environment dynamics by rule learning is crucial to our appealing results.

\begin{table}[t]
\caption{Ablation study of \ours with different configurations on Minecraft tasks, in the format of ``success rate (replanning rounds)''.
The success rate (\%) refers to the percentage of tasks completed successfully (higher the better). Replanning rounds (lower the better) measure the inference efficiency and represent the number of revisions needed for the agent to complete a task.
The row highlighted in grey represents the configuration and performance of \ours.}
\vspace{-0.5em}
\label{tab:minecraft_ablation_study}
\begin{center}
\renewcommand\arraystretch{1.1}
\resizebox{\linewidth}{!}{
    \begin{tabular}{llccccccc}
    \toprule
    \multicolumn{2}{l}{\textbf{\ours}} & \multicolumn{7}{c}{Success Rate (\%) $\uparrow$ (Replanning Rounds $\downarrow$)} \\ \cmidrule(lr){1-2} \cmidrule(lr){3-9}
    Agent         & World Model    & Avg.        & Wooden      & Stone      & Iron      & Golden    & Diamond   & Redstone    \\ 
    \midrule
    LLM           & -              & 37(35.36)   & 83(10.67)   & 41(33.26)  & 33(35.27) & 22(45.29) & 24(42.46) & 17(45.22)   \\
    LLM           & LLM            & 38(33.53)   & 86(10.35)   & 44(30.79)  & 35(34.08) & 19(43.99) & 26(39.51) & 19(42.46)    \\
    LLM+rules     & -              & 61(23.13)  & 94(5.04)    & 89(9.58)   & 67(18.56) & 33(39.67) & 41(32.73) & 43(33.21)     \\
    \rowcolor{gray!30} LLM & LLM+rules& 69(15.77)   & 98(1.64)    & 91(4.58)   & 63(19.38) & 69(15.61) & 46(27.08) & 48(26.33)     \\
    LLM+rules     & LLM+rules      & 67(16.59)   & 95(2.88)    & 93(3.75)   & 58(21.42) & 62(19.34) & 53(23.75) & 43(28.41)     \\ 
    \bottomrule
    \end{tabular}
    }
\end{center}
\end{table}

\section{Conclusion}

We have shown that LLMs can effectively serve as world models for agents when aligned with environment dynamics through rule learning. Our neurosymbolic approach bridges the gap between LLMs' prior knowledge and specific environments without gradient updates. By integrating a rule-enhanced LLM-based world model with MPC, our agent WALL-E demonstrates superior planning and task-solving abilities. Experiments indicate that WALL-E outperforms baselines in Minecraft and ALFWorld, achieving higher success rates with fewer replanning rounds and reduced token usage. Specifically, WALL-E attains a 15–30\% higher success rate in Minecraft, requires 8–20 fewer replanning rounds, and uses only 60–80\% of the tokens compared to baselines. In ALFWorld, it rapidly reaches a 95\% success rate from the 6th iteration onward. The rule learning converges swiftly by the 4th iteration, outperforming buffered trajectory methods in both efficiency and effectiveness. These results suggest that minimal additional rules suffice to align LLM predictions with environment dynamics, enhancing model-based agents in complex environments.

\bibliography{main_file}

\begin{thebibliography}{48}
\providecommand{\natexlab}[1]{#1}
\providecommand{\url}[1]{\texttt{#1}}
\expandafter\ifx\csname urlstyle\endcsname\relax
  \providecommand{\doi}[1]{doi: #1}\else
  \providecommand{\doi}{doi: \begingroup \urlstyle{rm}\Url}\fi

\bibitem[Brohan et~al.(2022)Brohan, Brown, Carbajal, Chebotar, Dabis, Finn, Gopalakrishnan, Hausman, Herzog, Hsu, et~al.]{rss/rt-1}
Anthony Brohan, Noah Brown, Justice Carbajal, Yevgen Chebotar, Joseph Dabis, Chelsea Finn, Keerthana Gopalakrishnan, Karol Hausman, Alex Herzog, Jasmine Hsu, et~al.
\newblock Rt-1: Robotics transformer for real-world control at scale.
\newblock \emph{arXiv preprint arXiv:2212.06817}, 2022.

\bibitem[Dai et~al.(2023)Dai, Li, Li, Tiong, Zhao, Wang, Li, Fung, and Hoi]{dai2023instructblip}
Wenliang Dai, Junnan Li, Dongxu Li, Anthony Tiong, Junqi Zhao, Weisheng Wang, Boyang Li, Pascale Fung, and Steven Hoi.
\newblock Instruct{BLIP}: Towards general-purpose vision-language models with instruction tuning.
\newblock In \emph{NeurIPS}, 2023.

\bibitem[Das et~al.(2018)Das, Datta, Gkioxari, Lee, Parikh, and Batra]{cvpr/EmbodiedQA}
Abhishek Das, Samyak Datta, Georgia Gkioxari, Stefan Lee, Devi Parikh, and Dhruv Batra.
\newblock Embodied question answering.
\newblock In \emph{CVPR}, 2018.

\bibitem[Fan et~al.(2022)Fan, Wang, Jiang, Mandlekar, Yang, Zhu, Tang, Huang, Zhu, and Anandkumar]{fan2022minedojo}
Linxi Fan, Guanzhi Wang, Yunfan Jiang, Ajay Mandlekar, Yuncong Yang, Haoyi Zhu, Andrew Tang, De-An Huang, Yuke Zhu, and Anima Anandkumar.
\newblock Minedojo: Building open-ended embodied agents with internet-scale knowledge.
\newblock \emph{NeurIPS}, 2022.

\bibitem[Gao et~al.(2023{\natexlab{a}})Gao, Sarkar, Xia, Xiao, Wu, Ichter, Majumdar, and Sadigh]{gao2023pg-vlm}
Jensen Gao, Bidipta Sarkar, Fei Xia, Ted Xiao, Jiajun Wu, Brian Ichter, Anirudha Majumdar, and Dorsa Sadigh.
\newblock Physically grounded vision-language models for robotic manipulation.
\newblock \emph{arXiv preprint arXiv:2309.02561}, 2023{\natexlab{a}}.

\bibitem[Gao et~al.(2023{\natexlab{b}})Gao, Han, Zhang, Lin, Geng, Zhou, Zhang, Lu, He, Yue, et~al.]{gao2023llama}
Peng Gao, Jiaming Han, Renrui Zhang, Ziyi Lin, Shijie Geng, Aojun Zhou, Wei Zhang, Pan Lu, Conghui He, Xiangyu Yue, et~al.
\newblock Llama-adapter v2: Parameter-efficient visual instruction model.
\newblock \emph{arXiv preprint arXiv:2304.15010}, 2023{\natexlab{b}}.

\bibitem[Ge et~al.(2024)Ge, Huang, Zhou, Li, Wang, Tang, and Zhuang]{ge2024worldgpt}
Zhiqi Ge, Hongzhe Huang, Mingze Zhou, Juncheng Li, Guoming Wang, Siliang Tang, and Yueting Zhuang.
\newblock Worldgpt: Empowering llm as multimodal world model.
\newblock \emph{arXiv preprint arXiv:2404.18202}, 2024.

\bibitem[Guan et~al.(2023)Guan, Valmeekam, Sreedharan, and Kambhampati]{guan2023leveraging}
Lin Guan, Karthik Valmeekam, Sarath Sreedharan, and Subbarao Kambhampati.
\newblock Leveraging pre-trained large language models to construct and utilize world models for model-based task planning.
\newblock \emph{NeurIPS}, 2023.

\bibitem[Hafner et~al.(2019)Hafner, Lillicrap, Ba, and Norouzi]{hafner2019dream}
Danijar Hafner, Timothy Lillicrap, Jimmy Ba, and Mohammad Norouzi.
\newblock Dream to control: Learning behaviors by latent imagination.
\newblock \emph{arXiv preprint arXiv:1912.01603}, 2019.

\bibitem[Hafner et~al.(2020)Hafner, Lillicrap, Norouzi, and Ba]{hafner2020mastering}
Danijar Hafner, Timothy Lillicrap, Mohammad Norouzi, and Jimmy Ba.
\newblock Mastering atari with discrete world models.
\newblock \emph{arXiv preprint arXiv:2010.02193}, 2020.

\bibitem[Hafner et~al.(2023)Hafner, Pasukonis, Ba, and Lillicrap]{hafner2023mastering}
Danijar Hafner, Jurgis Pasukonis, Jimmy Ba, and Timothy Lillicrap.
\newblock Mastering diverse domains through world models.
\newblock \emph{arXiv preprint arXiv:2301.04104}, 2023.

\bibitem[Hao et~al.(2023)Hao, Gu, Ma, Hong, Wang, Wang, and Hu]{hao2023reasoning}
Shibo Hao, Yi~Gu, Haodi Ma, Joshua~Jiahua Hong, Zhen Wang, Daisy~Zhe Wang, and Zhiting Hu.
\newblock Reasoning with language model is planning with world model.
\newblock \emph{arXiv preprint arXiv:2305.14992}, 2023.

\bibitem[Hu \& Shu(2023)Hu and Shu]{hu2023language}
Zhiting Hu and Tianmin Shu.
\newblock Language models, agent models, and world models: The law for machine reasoning and planning.
\newblock \emph{arXiv preprint arXiv:2312.05230}, 2023.

\bibitem[Li et~al.(2023)Li, Li, Savarese, and Hoi]{li2023blip}
Junnan Li, Dongxu Li, Silvio Savarese, and Steven Hoi.
\newblock Blip-2: Bootstrapping language-image pre-training with frozen image encoders and large language models.
\newblock In \emph{ICML}, 2023.

\bibitem[Li et~al.(2024{\natexlab{a}})Li, Chen, Wang, Nguyen, Li, and Zhou]{li2024ruler}
Ming Li, Han Chen, Chenguang Wang, Dang Nguyen, Dianqi Li, and Tianyi Zhou.
\newblock Ruler: Improving llm controllability by rule-based data recycling.
\newblock \emph{arXiv preprint arXiv:2406.15938}, 2024{\natexlab{a}}.

\bibitem[Li et~al.(2024{\natexlab{b}})Li, Xie, Shao, Chen, Jiang, and Nie]{li2024optimus}
Zaijing Li, Yuquan Xie, Rui Shao, Gongwei Chen, Dongmei Jiang, and Liqiang Nie.
\newblock Optimus-1: Hybrid multimodal memory empowered agents excel in long-horizon tasks.
\newblock \emph{arXiv preprint arXiv:2408.03615}, 2024{\natexlab{b}}.

\bibitem[Liu et~al.(2024)Liu, Chen, Bai, Luo, Song, Jiang, Li, Zhao, Lin, Li, et~al.]{liu2024aligning}
Yang Liu, Weixing Chen, Yongjie Bai, Jingzhou Luo, Xinshuai Song, Kaixuan Jiang, Zhida Li, Ganlong Zhao, Junyi Lin, Guanbin Li, et~al.
\newblock Aligning cyber space with physical world: A comprehensive survey on embodied ai.
\newblock \emph{arXiv preprint arXiv:2407.06886}, 2024.

\bibitem[Liu et~al.(2023)Liu, Hu, Zhang, Guo, Ke, Liu, and Wang]{liu2023reason}
Zhihan Liu, Hao Hu, Shenao Zhang, Hongyi Guo, Shuqi Ke, Boyi Liu, and Zhaoran Wang.
\newblock Reason for future, act for now: A principled framework for autonomous llm agents with provable sample efficiency.
\newblock \emph{arXiv preprint arXiv:2309.17382}, 2023.

\bibitem[Luo et~al.(2023)Luo, Ju, Xiong, Li, Haffari, and Pan]{luo2023chatrule}
Linhao Luo, Jiaxin Ju, Bo~Xiong, Yuan-Fang Li, Gholamreza Haffari, and Shirui Pan.
\newblock Chatrule: Mining logical rules with large language models for knowledge graph reasoning.
\newblock \emph{arXiv preprint arXiv:2309.01538}, 2023.

\bibitem[Micheli \& Fleuret(2021)Micheli and Fleuret]{micheli2021language}
Vincent Micheli and Fran{\c{c}}ois Fleuret.
\newblock Language models are few-shot butlers.
\newblock \emph{arXiv preprint arXiv:2104.07972}, 2021.

\bibitem[Mu et~al.(2023{\natexlab{a}})Mu, Chen, Wang, Chen, Karamardian, Aljeraisy, Hendrycks, and Wagner]{mu2023can}
Norman Mu, Sarah Chen, Zifan Wang, Sizhe Chen, David Karamardian, Lulwa Aljeraisy, Dan Hendrycks, and David Wagner.
\newblock Can llms follow simple rules?
\newblock \emph{arXiv preprint arXiv:2311.04235}, 2023{\natexlab{a}}.

\bibitem[Mu et~al.(2023{\natexlab{b}})Mu, Zhang, Hu, Wang, Ding, Jin, Wang, Dai, Qiao, and Luo]{corr/embodiedgpt}
Yao Mu, Qinglong Zhang, Mengkang Hu, Wenhai Wang, Mingyu Ding, Jun Jin, Bin Wang, Jifeng Dai, Yu~Qiao, and Ping Luo.
\newblock Embodiedgpt: Vision-language pre-training via embodied chain of thought.
\newblock \emph{arXiv preprint arXiv:2305.15021}, 2023{\natexlab{b}}.

\bibitem[OpenAI(2023)]{gpt-4}
OpenAI.
\newblock {GPT-4} technical report.
\newblock \emph{arXiv preprint arXiv.2303.08774}, 2023.

\bibitem[Qin \& Badgwell(2003)Qin and Badgwell]{qin2003survey}
S~Joe Qin and Thomas~A Badgwell.
\newblock A survey of industrial model predictive control technology.
\newblock \emph{Control engineering practice}, 11\penalty0 (7):\penalty0 733--764, 2003.

\bibitem[Rafailov et~al.(2024)Rafailov, Sharma, Mitchell, Manning, Ermon, and Finn]{rafailov2024direct}
Rafael Rafailov, Archit Sharma, Eric Mitchell, Christopher~D Manning, Stefano Ermon, and Chelsea Finn.
\newblock Direct preference optimization: Your language model is secretly a reward model.
\newblock \emph{NeurIPS}, 2024.

\bibitem[Shinn et~al.(2024)Shinn, Cassano, Gopinath, Narasimhan, and Yao]{shinn2024reflexion}
Noah Shinn, Federico Cassano, Ashwin Gopinath, Karthik Narasimhan, and Shunyu Yao.
\newblock Reflexion: Language agents with verbal reinforcement learning.
\newblock \emph{NeurIPS}, 2024.

\bibitem[Shridhar et~al.(2020{\natexlab{a}})Shridhar, Thomason, Gordon, Bisk, Han, Mottaghi, Zettlemoyer, and Fox]{cvpr/alfred}
Mohit Shridhar, Jesse Thomason, Daniel Gordon, Yonatan Bisk, Winson Han, Roozbeh Mottaghi, Luke Zettlemoyer, and Dieter Fox.
\newblock {ALFRED:} {A} benchmark for interpreting grounded instructions for everyday tasks.
\newblock In \emph{CVPR}, 2020{\natexlab{a}}.

\bibitem[Shridhar et~al.(2020{\natexlab{b}})Shridhar, Yuan, C{\^o}t{\'e}, Bisk, Trischler, and Hausknecht]{shridhar2020alfworld}
Mohit Shridhar, Xingdi Yuan, Marc-Alexandre C{\^o}t{\'e}, Yonatan Bisk, Adam Trischler, and Matthew Hausknecht.
\newblock Alfworld: Aligning text and embodied environments for interactive learning.
\newblock \emph{arXiv preprint arXiv:2010.03768}, 2020{\natexlab{b}}.

\bibitem[Sun et~al.(2024)Sun, Zhuang, Kong, Dai, and Zhang]{sun2024adaplanner}
Haotian Sun, Yuchen Zhuang, Lingkai Kong, Bo~Dai, and Chao Zhang.
\newblock Adaplanner: Adaptive planning from feedback with language models.
\newblock \emph{NeurIPS}, 2024.

\bibitem[Tang et~al.(2024)Tang, Key, and Ellis]{tang2024worldcoder}
Hao Tang, Darren Key, and Kevin Ellis.
\newblock Worldcoder, a model-based llm agent: Building world models by writing code and interacting with the environment.
\newblock \emph{arXiv preprint arXiv:2402.12275}, 2024.

\bibitem[Wang et~al.(2023{\natexlab{a}})Wang, Cai, Chen, Liu, Ma, Liang, and CraftJarvis]{wang2023describe}
Zihao Wang, Shaofei Cai, Guanzhou Chen, Anji Liu, Xiaojian Ma, Yitao Liang, and Team CraftJarvis.
\newblock Describe, explain, plan and select: interactive planning with large language models enables open-world multi-task agents.
\newblock In \emph{NeurIPS}, 2023{\natexlab{a}}.

\bibitem[Wang et~al.(2023{\natexlab{b}})Wang, Cai, Liu, Jin, Hou, Zhang, Lin, He, Zheng, Yang, et~al.]{wang2023jarvis}
Zihao Wang, Shaofei Cai, Anji Liu, Yonggang Jin, Jinbing Hou, Bowei Zhang, Haowei Lin, Zhaofeng He, Zilong Zheng, Yaodong Yang, et~al.
\newblock Jarvis-1: Open-world multi-task agents with memory-augmented multimodal language models.
\newblock \emph{arXiv preprint arXiv:2311.05997}, 2023{\natexlab{b}}.

\bibitem[Wei et~al.(2022)Wei, Wang, Schuurmans, Bosma, Ichter, Xia, Chi, Le, and Zhou]{nips/cot}
Jason Wei, Xuezhi Wang, Dale Schuurmans, Maarten Bosma, Brian Ichter, Fei Xia, Ed~H. Chi, Quoc~V. Le, and Denny Zhou.
\newblock Chain-of-thought prompting elicits reasoning in large language models.
\newblock In \emph{NeurIPS}, 2022.

\bibitem[Wong et~al.(2023)Wong, Mao, Sharma, Siegel, Feng, Korneev, Tenenbaum, and Andreas]{wong2023learning}
Lionel Wong, Jiayuan Mao, Pratyusha Sharma, Zachary~S Siegel, Jiahai Feng, Noa Korneev, Joshua~B Tenenbaum, and Jacob Andreas.
\newblock Learning adaptive planning representations with natural language guidance.
\newblock \emph{arXiv preprint arXiv:2312.08566}, 2023.

\bibitem[Wu et~al.(2023)Wu, Bansal, Zhang, Wu, Zhang, Zhu, Li, Jiang, Zhang, and Wang]{wu2023autogen}
Qingyun Wu, Gagan Bansal, Jieyu Zhang, Yiran Wu, Shaokun Zhang, Erkang Zhu, Beibin Li, Li~Jiang, Xiaoyun Zhang, and Chi Wang.
\newblock Autogen: Enabling next-gen llm applications via multi-agent conversation framework.
\newblock \emph{arXiv preprint arXiv:2308.08155}, 2023.

\bibitem[Wu et~al.(2024)Wu, Guan, Li, Huang, Liu, Wang, Xian, Shrivastava, Huang, Boyd-Graber, et~al.]{wu2024autohallusion}
Xiyang Wu, Tianrui Guan, Dianqi Li, Shuaiyi Huang, Xiaoyu Liu, Xijun Wang, Ruiqi Xian, Abhinav Shrivastava, Furong Huang, Jordan~Lee Boyd-Graber, et~al.
\newblock Autohallusion: Automatic generation of hallucination benchmarks for vision-language models.
\newblock \emph{arXiv preprint arXiv:2406.10900}, 2024.

\bibitem[Xiang et~al.(2024)Xiang, Tao, Gu, Shu, Wang, Yang, and Hu]{xiang2024language}
Jiannan Xiang, Tianhua Tao, Yi~Gu, Tianmin Shu, Zirui Wang, Zichao Yang, and Zhiting Hu.
\newblock Language models meet world models: Embodied experiences enhance language models.
\newblock \emph{NeurIPS}, 2024.

\bibitem[Xiao et~al.(2021)Xiao, Mehdipour, Collin, Bin-Nun, Frazzoli, Tebbens, and Belta]{xiao2021rule}
Wei Xiao, Noushin Mehdipour, Anne Collin, Amitai~Y Bin-Nun, Emilio Frazzoli, Radboud~Duintjer Tebbens, and Calin Belta.
\newblock Rule-based optimal control for autonomous driving.
\newblock In \emph{Proceedings of the ACM/IEEE 12th International Conference on Cyber-Physical Systems}, pp.\  143--154, 2021.

\bibitem[Xie et~al.(2024)Xie, Yang, Gunerli, and Riedl]{xie2024making}
Kaige Xie, Ian Yang, John Gunerli, and Mark Riedl.
\newblock Making large language models into world models with precondition and effect knowledge.
\newblock \emph{arXiv preprint arXiv:2409.12278}, 2024.

\bibitem[Yang et~al.(2023{\natexlab{a}})Yang, Lin, Zhou, and Wen]{yang2023enabling}
Wenkai Yang, Yankai Lin, Jie Zhou, and Jirong Wen.
\newblock Enabling large language models to learn from rules.
\newblock \emph{arXiv preprint arXiv:2311.08883}, 2023{\natexlab{a}}.

\bibitem[Yang et~al.(2024)Yang, Zhou, Li, Tao, Li, Shen, He, Jiang, and Shi]{yang2024embodied}
Yijun Yang, Tianyi Zhou, Kanxue Li, Dapeng Tao, Lusong Li, Li~Shen, Xiaodong He, Jing Jiang, and Yuhui Shi.
\newblock Embodied multi-modal agent trained by an llm from a parallel textworld.
\newblock In \emph{CVPR}, 2024.

\bibitem[Yang et~al.(2023{\natexlab{b}})Yang, Li, and Liu]{yang2023failures}
Zeyuan Yang, Peng Li, and Yang Liu.
\newblock Failures pave the way: Enhancing large language models through tuning-free rule accumulation.
\newblock \emph{arXiv preprint arXiv:2310.15746}, 2023{\natexlab{b}}.

\bibitem[Yao et~al.(2023)Yao, Zhao, Yu, Du, Shafran, Narasimhan, and Cao]{yao2023react}
Shunyu Yao, Jeffrey Zhao, Dian Yu, Nan Du, Izhak Shafran, Karthik Narasimhan, and Yuan Cao.
\newblock React: Synergizing reasoning and acting in language models.
\newblock In \emph{ICLR}, 2023.

\bibitem[Zhao et~al.(2024)Zhao, Lee, and Hsu]{zhao2024large}
Zirui Zhao, Wee~Sun Lee, and David Hsu.
\newblock Large language models as commonsense knowledge for large-scale task planning.
\newblock \emph{NeurIPS}, 2024.

\bibitem[Zhu et~al.(2023{\natexlab{a}})Zhu, Chen, Shen, Li, and Elhoseiny]{zhu2023minigpt}
Deyao Zhu, Jun Chen, Xiaoqian Shen, Xiang Li, and Mohamed Elhoseiny.
\newblock Minigpt-4: Enhancing vision-language understanding with advanced large language models.
\newblock \emph{arXiv preprint arXiv:2304.10592}, 2023{\natexlab{a}}.

\bibitem[Zhu et~al.(2023{\natexlab{b}})Zhu, Chen, Tian, Tao, Su, Yang, Huang, Li, Lu, Wang, et~al.]{zhu2023ghost}
Xizhou Zhu, Yuntao Chen, Hao Tian, Chenxin Tao, Weijie Su, Chenyu Yang, Gao Huang, Bin Li, Lewei Lu, Xiaogang Wang, et~al.
\newblock Ghost in the minecraft: Generally capable agents for open-world environments via large language models with text-based knowledge and memory.
\newblock \emph{arXiv preprint arXiv:2305.17144}, 2023{\natexlab{b}}.

\bibitem[Zhu et~al.(2023{\natexlab{c}})Zhu, Xue, Chen, Zhou, Tang, Schuurmans, and Dai]{Zhu2023Large}
Zhaocheng Zhu, Yuan Xue, Xinyun Chen, Denny Zhou, Jian Tang, D.~Schuurmans, and Hanjun Dai.
\newblock Large language models can learn rules.
\newblock \emph{arXiv preprint arXiv:2310.07064}, 2023{\natexlab{c}}.

\bibitem[Zitkovich et~al.(2023)Zitkovich, Yu, Xu, Xu, Xiao, Xia, Wu, Wohlhart, Welker, Wahid, et~al.]{corr/rt-2}
Brianna Zitkovich, Tianhe Yu, Sichun Xu, Peng Xu, Ted Xiao, Fei Xia, Jialin Wu, Paul Wohlhart, Stefan Welker, Ayzaan Wahid, et~al.
\newblock Rt-2: Vision-language-action models transfer web knowledge to robotic control.
\newblock In \emph{CoRL}, 2023.

\end{thebibliography}
\bibliographystyle{paper_conference}

\appendix

\newpage

\section{Detailed Related Work} 
\label{sec:Detailed Related Work}

\paragraph{LLMs with Rule Learning.}
Recent studies have explored integrating LLMs with rule learning to enhance reasoning and model behavior. For instance, \cite{yang2023enabling} introduced rule distillation, enabling LLMs to learn from predefined rules, which improved generalization with limited training data. Similarly, \cite{Zhu2023Large} proposed the Hypotheses-to-Theories (HtT) framework, which enhanced numerical and relational reasoning by generating and validating rules from training data. In the same vein, \cite{mu2023can} developed the RuLES framework to evaluate LLM adherence to developer-specified rules, addressing challenges like rule evasion through adversarial inputs. Furthermore, \cite{yang2023failures} presented the Tuning-free Rule Accumulation (TRAN) framework, allowing LLMs to accumulate rules from incorrect cases to avoid repeating mistakes without additional tuning. Lastly, in knowledge graph reasoning, \cite{luo2023chatrule} introduced ChatRule, a framework that mines logical rules over knowledge graphs using LLMs.

These studies show the potential of combining LLMs with rule learning to improve reasoning and generalization. However, none have integrated rule learning with LLM-based world models, which is the focus of our work. We explore how rule learning can align LLM world models with specific environment dynamics, thereby improving the performance of model-based agents in dynamic environments.

\paragraph{Using LLMs to Build World Models.}
Many studies have leveraged LLMs to construct world models for planning. For example, \cite{wong2023learning} proposed translating natural language instructions into adaptable planning representations via LLMs, enabling flexible and context-aware world modeling. Similarly, \cite{guan2023leveraging} showed that combining pre-trained LLMs with task-specific planning modules improves task success rates by providing a more detailed understanding of the environment. Another approach, WorldCoder \cite{tang2024worldcoder}, exemplified an LLM agent that constructs world models by generating and executing code to simulate various states and actions, refining its understanding iteratively.

These studies demonstrate the utility of LLMs in building world models to improve planning and reasoning in complex environments. However, unlike these works, our approach directly employs the LLM as the world model, utilizing its inherent knowledge and reasoning abilities without an explicit model-building phase. This direct use of LLMs enhances adaptability and computational efficiency.

\paragraph{Using LLMs as World Models.}
Several studies have explored using LLMs directly as world models by leveraging their implicit knowledge. Some methods rely on fine-tuning to align the LLM world model with the environment. For example, \cite{xiang2024language} fine-tuned LLMs with embodied experiences in a simulated world to enhance reasoning and planning abilities in embodied environments. Similarly, \cite{xie2024making} transformed LLMs into world models by incorporating knowledge of action preconditions and effects, fine-tuning the models to reason about actions and predict their outcomes accurately.

Other approaches align LLMs as world models through prompting. For instance, \cite{zhao2024large} introduced the LLM-MCTS algorithm, prompting LLMs to serve as both the policy and world model for large-scale task planning, integrating commonsense priors with guided search. In another approach, \cite{hao2023reasoning} introduced Reasoning via Planning (RAP), where LLMs are prompted to act as reasoning agents and world models by generating reasoning trees to explore solutions. Finally, \citep{liu2023reason} used a Bayesian adaptive Markov Decision Process to guide LLMs in planning future trajectories, prompting them to predict future states.

While these approaches demonstrate the potential of using LLMs as world models, they often require extensive fine-tuning or rely heavily on human-crafted prompts, making them labor-intensive and inflexible. Our work overcomes these limitations by automatically extracting rules from exploration experiences, reducing human effort and enhancing adaptability across different environments.

\section{Detailed Prompt}

\subsection{Learn New Rules from Real Trajectories}
\label{sec:Prompt for Learn New Rules from Real Trajectories}

\begin{center}
    \textbf{Prompt for Learning New Rules from Real Trajectories}
\end{center}

\lstset{language=}
\begin{lstlisting}
You are responsible for mining new rules from the given transitions, ensuring that these rules differ from the ones already provided. 
Focus on generating general and universal rules that are not tied to any specific item or tool. 
Your goal is to generalize across different objects, creating flexible rules that can be applied broadly to diverse contexts and situations.

I will give you an array of transitions:
[
    {
        'state_0': {
            "state feature 1": {"feature name": value, ...},
            ...
        }, 
        'action': {
            "name": "action name", 
            "action feature 1": {"feature name": value, ...},
            ...
        }, 
        'action_result': {
        "feedback": "the environment feedback",
        "success": "Whether the action is executed successfully, give 'True' or 'False' only",
        "suggestion": "If the 'action' fails, 'suggestion' would be given based on 'state 0' and 'action'"
    }
    },
    {
        'state_0': {
            "state feature 1": {"feature name": value, ...},
            ...
        }, 
        'action': {
            "name": "action name", 
            "action feature 1": {"feature name": value, ...},
            ...
        }, 
        'action_result': {
        "feedback": "the environment feedback",
        "success": "Whether the action is executed successfully, give 'True' or 'False' only",
        "suggestion": "If the 'action' fails, 'suggestion' would be given based on 'state 0' and 'action'"
    }
    },
    ...
]
and an array of rules:
[
    "Rule 1: For action ..., if..., the action will fail; Checking Method: ...",
    "Rule 2: For action ..., if..., the action will fail; Checking Method: ...",
    ...
]

You should only respond in the format as described below:
RESPONSE FORMAT:
{
    "new_rules":[
        "Rule ...: For action ...,...; Checking Method: ...",
        "Rule ...: For action ...,...; Checking Method: ...",
        ...
    ]
}

Instructions:
- Ensure the response can be parsed by Python `json.loads`, e.g.: no trailing commas, **no single quotes**, etc.
- Please use you knowledge in <ENV>, do inductive reasoning. You need to dig up as many rules as possible that satisfy all transitions.
- Extract and utilize only the features that influence the outcome of the action.
- Please generate general and universal rules; the rules should not reference any specific item or tool! You need to generalize across various items or tools.
- Generate only the rules under what conditions the action will fail.
- While generating a rule, you also need to state how to check if a transition satisfies this rule. Please be specific as to which and how 'features' need to be checked
\end{lstlisting}

\subsection{Refine Learned Rules}
\label{sec:Refine Learned Rules}

\begin{center}
    \textbf{Prompt for Refining Learned Rules}
\end{center}

\lstset{language=}
\begin{lstlisting}
You are responsible for improving the existing rules by verifying that they hold true for all transitions. 
This involves identifying any conflicting rules, diagnosing potential issues, and making necessary modifications. 
Ensure that the refined rules are consistent and correctly align with the transitions provided, avoiding any contradictions or overlaps.

I will give you an array of transitions:
[
    {
        'state_0': {
            "state feature 1": {"feature name": value, ...},
            ...
        }, 
        'action': {
            "name": "action name", 
            "action feature 1": {"feature name": value, ...},
            ...
        }, 
        'action_result': {
        "feedback": "the environment feedback",
        "success": "Whether the action is executed successfully, give 'True' or 'False' only",
        "suggestion": "If the 'action' fails, 'suggestion' would be given based on 'state 0' and 'action'"
    }
    },
    {
        'state_0': {
            "state feature 1": {"feature name": value, ...},
            ...
        }, 
        'action': {
            "name": "action name", 
            "action feature 1": {"feature name": value, ...},
            ...
        }, 
        'action_result': {
        "feedback": "the environment feedback",
        "success": "Whether the action is executed successfully, give 'True' or 'False' only",
        "suggestion": "If the 'action' fails, 'suggestion' would be given based on 'state 0' and 'action'"
    }
    },
    ...
]
and an array of rules:
[
    "Rule 1: For action ..., if..., the action will fail; Checking Method: ...",
    "Rule 2: For action ..., if..., the action will fail; Checking Method: ...",
    ...
]

You should only respond in the format as described below:
RESPONSE FORMAT:
{
    "verified_rules":[
        "Rule ...: For action ...,...; Checking Method: ...",
        "Rule ...: For action ...,...; Checking Method: ...",
        ...
    ],
    "conflicting_rules":[
        "Rule ...: For action ...,...; Checking Method: ...",
        "Rule ...: For action ...,...; Checking Method: ...",
        ...
    ],
    "improved_rules":[
        "Rule ...: For action ...,...; Checking Method: ...",
        "Rule ...: For action ...,...; Checking Method: ...",
        ...
    ],
    "final_rules":[
        "Rule ...: For action ...,...; Checking Method: ...",
        "Rule ...: For action ...,...; Checking Method: ...",
        ...
    ]
}

where
verified_rules: list rules that satisfy all the provided transitions.
conflicting_rules: list rules that contradict any of the transitions. Modify these rules if they can be modified correctly and put them in 'improved_rules'.
improved_rules: show modified 'conflicting_rules'.
final_rules: combine all the rules from 'verified_rules', 'new_rules'.


Instructions:
- Ensure the response can be parsed by Python `json.loads`, e.g.: no trailing commas, **no single quotes**, etc.
- Please use you knowledge in <ENV>, do inductive reasoning. You need to dig up as many rules as possible that satisfy all transitions.
- Extract and utilize only the features that influence the outcome of the action.
- Please generate general and universal rules; the rules should not reference any specific item or tool! You need to generalize across various items or tools.
- Generate only the rules under what conditions the action will fail.
- While generating a rule, you also need to state how to check if a transition satisfies this rule. Please be specific as to which and how 'features' need to be checked
\end{lstlisting}

\subsection{Translate Natural Language Rules to Code }
\label{sec:Translate Natural Language Rules to Code}

\begin{center}
    \textbf{Prompt for Translating Natural Language Rules to Code}
\end{center}

\lstset{language=}
\begin{lstlisting}
You are responsible for generating code rules by implementing the learned rules in Python. 
Your task is to write a function that takes the current state and an action as inputs, evaluates these conditions, and returns a Boolean value based on the specified rule. 
This function should effectively mirror the logic of the rules, enabling precise predictions for various state-action pairs.

The function should be defined as follows:

```python
def expected_rule_code(state, action):
    # Your code here
    return feedback, success, suggestion
where
feedback: a string, give the action feedback based on success or not.
success: a bool, whether the action is executed successfully, give 'True' or 'False'. If the action type is not the action type in the rule, count as success (e.g., success = True).
suggestion: a string, if the 'action' fails, 'suggestion' would be given based on 'rule', 'state' and 'action'.

Here is several examples of the input format:
<Input Format>

The function should return a Boolean (True or False) based on an internal rule which you must implement.

Ensure that the function handles the input and outputs the expected result based on <ENV>'s mechanics and the provided state and action.

If the rule involves the need to use your knowledge to make a judgement about an item or action then write the function, LLM_request("question"+"response format"). 
LLM_request would send the "question" to gpt4, and return the gpt4's response. you just need to write the "question" in the LLM_request. 
LLM_request("question"+"response format") has already been predefined, you can just use it dirtectly. Do not need to define it again in your response. But you need to define the "question" and "response format" carefully.

example: i want to know if the item can be destroyed
the LLM function: LLM_request(f"if the {item} can be destroyed in the <ENV>?" + "only reply True or False")

You should only respond in the format as described below, and do not give example usage or anything else:
RESPONSE FORMAT:
def expected_rule_code(state, action):
    # Your code here
\end{lstlisting}
where ``input format'' please refer to Appendix \ref{sec: Environments' State Space and Action Space}.

\section{Environments' State Space and Action Space}
\label{sec: Environments' State Space and Action Space}

The format of state and action information is crucial for understanding the rules we have extracted. In this section, we provide an description of the state and action space used in different environments.

\subsection{Minecraft}

\textbf{State Space.} We collect state information directly from the observation space provided by MineDojo~\citep{fan2022minedojo}, which includes: (1) equipment status, (2) inventory details, (3) life statistics, and (4) location statistics. The specific structure is illustrated in the following example.

\begin{center}
    \textbf{Examples for Minecraft's State Space}
\end{center}
\lstset{language=}
\begin{lstlisting}
state = {
    "equipment": {
        "dirt": 60.0,
        "diamond boots": 1.0,
        "diamond leggings": 1.0,
        "diamond chestplate": 1.0,
        "diamond helmet": 1.0,
        "air": 0.0
    },
    "inventory": {
        "dirt": 60.0,
        "crafting table": 1.0,
        "planks": 2.0,
        "stick": 4.0,
        "air": 0.0,
        "log": 1.0
    },
    "life_stats": {
        "life": 20.0,
        "oxygen": 300.0,
        "armor": 20.0,
        "food": 20.0,
        "saturation": 5.0,
        "is_sleeping": False
    },
    "location_stats": {
        "biome": "plains",
        "rainfall": 0.4,
        "temperature": 0.8,
        "is_raining": False,
        "sky_light_level": 0.2,
        "sun_brightness": 0.0
    }
}
\end{lstlisting}

\textbf{Action Space.} The action space is defined based on the action API provided by MineDojo~\citep{fan2022minedojo}, with additional modifications inspired by the action space used in GITM~\citep{zhu2023ghost}. The detailed action definitions are presented below.

\begin{center}
    \textbf{Minecraft's Action Space}
\end{center}
\lstset{language=}
\begin{lstlisting}
craft(obj, materials, platform): craft the object with the materials and platform; used to craft new object that is not in the inventory or is not enough.
- obj: a dict, whose key is the name of the object and value is the object quantity, like {"crafting table": 1} and {"stone pickaxe": 1}.
- materials: a dict, whose keys are the names of the materials and values are the quantities, like {"planks": 4} and {"cobblestone": 3, "stick": 2}.
- platform: a string, the platform used for crafting the current 'object', like "furnace" and "crafting table". Set to null if without any platform.

mine(obj, tool, y_level): dig down to the y-level and mine the specified object with the tool. This action will go underground and continuously mine the object until the desired quantity is obtained.
- obj: a dict, whose key is the name of the object and value is the object quantity, like {"stone": 5} and {"iron ore": 1}.
- tool (string): the tool used for mining, like "wooden pickaxe". Set to null if without any tool.
- y_level: a number, the y-level to dig down to. Different ores have different probabilities of distribution in different levels.

fight(obj, target, tool): find, track, and fight the target until you collect the desired number (goal_num) of object by using the chosen tool.
- obj: a dict, whose key is the name of the object and value is the object quantity, like {"leather": 5} and {"porkchop": 3}.
- target: a string, The name of the entity you want to fight (e.g., "skeleton", "sheep").
- tool: a string, the tool or weapon you will use in the fight, like "iron sword" or "wooden sword". Set to null if without any tool.

equip(obj): equip the object from the inventory.
- obj: a string, the object to equip, like "wooden pickaxe".

apply(obj, target, tool): automates the process of using a tool on target until you collect a specific number of object.
- obj: a dict, whose key is the name of the object and value is the object quantity, like {"wool": 5}.
- target: a string, the name of the target you want to interact with (e.g., "water", "sheep").
- tool: a string, the specific tool you will use for the action. (e.g., "bucket", "shears")

gather(obj, tool): collect resources (obj) directly from the environment. This includes picking up flowers, seeds from grass, and wood from trees.
- obj: a dict, whose key is the name of the object and value is the object quantity, like {"log": 10}.
- tool: a string, the tool you will use in the gathering. Set to null if without any tool.

change_time(target_time): adjust to the specified time of day; this function enables you to wait until a predefined time, such as morning, night, or midnight, depending on the specified target_time.
- target_time:  a string, specifying the desired time to change to. Valid options include "morning", "night", and "midnight", each corresponding to distinct values in 'sky_light_level' and 'sun_brightness' in "state features" like:
-- "morning": 'sky_light_level': array([1.]), 'sun_brightness': array([1.])
-- "night": 'sky_light_level': array([0.25]), 'sun_brightness': array([0.36])
-- "midnight": 'sky_light_level': array([0.2]), 'sun_brightness': array([0.])
\end{lstlisting}

\subsection{ALFWorld}
\label{sec: ALFworld Environments' State Space and Action Space}

\textbf{State Space.} In the original ALFWorld setup, state information is represented as natural language dialogue history. To facilitate the rule learning process, we developed scripts to transform this dialogue history into a structured JSON format, as shown in the following example.

\begin{center}
    \textbf{Examples for ALFWorld's State Space}
\end{center}
\lstset{language=}
\begin{lstlisting}
state = {
    "reachable_locations": [
        "cabinet 5",
        "cabinet 4",
        "cabinet 3",
        "cabinet 2",
        "cabinet 1",
        "coffeemachine 1",
        "countertop 2",
        "countertop 1",
        "diningtable 1",
        "drawer 2",
        "drawer 1",
        "fridge 1",
        "garbagecan 1",
        "microwave 1",
        "shelf 3",
        "shelf 2",
        "shelf 1",
        "sinkbasin 1",
        "stoveburner 4",
        "stoveburner 3",
        "stoveburner 2",
        "stoveburner 1",
        "toaster 1"
    ],
    "items_in_locations": {
        "fridge 1": [
            "lettuce 2",
            "mug 2",
            "potato 3"
        ],
        "microwave 1": []
    },
    "item_in_hand": {
        "item_name": "cup 1",
        "status": "normal"
    },
    "current_position": {
        "location_name": "microwave 1",
        "status": "open"
    }
}
\end{lstlisting}

\textbf{Action Space.} We utilize the action space provided by the ALFWorld directly, as demonstrated below. 
\begin{center}
    \textbf{Action Space for Minecraft}
\end{center}
\lstset{language=}
\begin{lstlisting}
go to [location/object]: Move to a specified location or object. 
open [object]: Open a specified object like a cabinet or drawer. 
close [object]: Close an opened object.
take [object] from [location]: Pick up an item from a specified location.
put [object] in/on [location]: Place an item in or on a specified location.
clean [object] with [location/tool]: Clean an object using a specific location or tool, like cleaning lettuce at the sink basin.
heat [object] with [tool]: Use an appliance, such as a microwave, to heat an item.
cool [object] with [tool]: Use a cooling tool or appliance, such as a fridge, to cool an item.
use [tool]: Activate or use a tool, such as a desklamp.
\end{lstlisting}

\section{Learned Rules}
\label{sec:Learned Rules}

There are two points to note about the numbering of the rules:
\begin{itemize}
    \item The reason for duplicates is that the numbering is based on actions, and different actions have their own separate sequences. For example: Rules for Craft: [Rule 1, Rule 2, Rule 3, Rule 4, Rule 5...]; Rules for Mine: [Rule 1, Rule 2, Rule 3, Rule 4, Rule 5...].
    \item The reason the sequence may appear unordered is that some rules have been pruned (Section \ref{sec:Rule Set Pruning via Maximum Coverage} Rule Set Pruning via Maximum Coverage). For instance, Rules for Craft where [Rule 1, Rule 2, Rule 4, Rule 5] has been removed, Rules for Mine where [Rule 1, Rule 3, Rule 4, Rule 5, Rule 6] has been removed, and the final rule set is Rules for Craft: [Rule 3, Rule 6] and Rules for Mine: [Rule 2, Rule 7].
\end{itemize}

\subsection{Natural Language Rules}

\begin{center}
    \textbf{Natural Language Rules for Minecraft}
\end{center}

\lstset{language=}
\begin{lstlisting}
"Rule 3: For action 'craft', if the specified platform is incorrect or not specified when required, the action will fail; Checking Method: Check if the 'platform' specified in the 'action' matches the required platform for the 'obj' being crafted.",
"Rule 6: For action 'craft', if the player does not have enough materials to craft the specified object, the action will fail; Checking Method: Check if the 'materials' specified in the 'action' are present in the 'inventory' with the required quantities. If not, the action will fail.",
"Rule 2: For action 'mine', if the 'tool' is not appropriate for the object being mined, the action will fail; Checking Method: Check if 'action.args.tool' is not suitable for 'action.args.obj'.",
"Rule 7: For action 'mine', if the 'tool' is not in the inventory, the action will fail; Checking Method: Check if 'action.args.tool' is not present in 'state_0.inventory'.",
"Rule 2: For action 'gather', if the 'sky_light_level' in 'location_stats' is less than 1.0, the action will fail; Checking Method: Check if 'sky_light_level' in 'location_stats' is less than 1.0.",
"Rule 1: For action 'fight', if the 'tool' is not present in the 'inventory' or 'equipment', the action will fail; Checking Method: Check if the 'tool' specified in the action is present in either 'inventory' or 'equipment'.",
\end{lstlisting}

\begin{center}
    \textbf{Natural Language Rules for ALFWorld}
\end{center}

\lstset{language=}
\begin{lstlisting}
Rule 1: For action 'clean', if the object to be cleaned is not in hand, the action will fail; Checking Method: Check if 'item_in_hand.item_name' in 'inital_state' matches 'action.args.obj'. 
Rule 3: For action clean, if the tool is not reachable, the action will fail; Checking Method: Check if the tool specified in the action is in the list of reachable locations in the initial state. 
Rule 5: For action 'clean', if the current position is not at the tool location, the action will fail; Checking Method: Check if 'current_position.location_name' in 'inital_state' matches 'action.args.tool'. 
Rule 2: For action 'take', if the agent is already holding an item, the action will fail; Checking Method: Check if 'item_in_hand.item_name' in 'inital_state' is not None. 
Rule 4: For action 'take', if the agent is not at the location of the item, the action will fail; Checking Method: Check if the 'current_position.location_name' in 'inital_state' is not the same as the 'source' in the 'action'. 
Rule 3: For action 'open', if the current position is not the target, the action will fail; Checking Method: Check if the 'current_position' is the target. 
Rule 2: For action 'put', if the item to be put is not in hand, the action will fail; Checking Method: Check if 'item_in_hand.item_name' is not equal to 'action.args.obj'. 
Rule 1: For action 'use', if the object to be used is not at the current position, the action will fail; Checking Method: Check if the object specified in the action is listed under the 'items_in_locations' of the 'current_position' in the 'inital_state'. 
Rule 1: For action 'heat', if the tool (microwave) is not at the current position, the action will fail; Checking Method: Check if 'current_position.location_name' is equal to the tool in the action arguments. 
Rule 5: For action 'heat', if the item in hand is not the item to be heated, the action will fail; Checking Method: Check if 'item_in_hand' in 'inital_state' is equal to 'action.args.obj'. 
Rule 1: For action 'go to', if the target location is the same as the current location, the action will fail; Checking Method: Check if 'current_position.location_name' is equal to 'action.args.target'. 
\end{lstlisting}

\subsection{Code-based Rules}
\label{sec:Code-based Rules}

When a rule requires the LLM's domain knowledge to make judgments, we instruct the LLM to use the function $\texttt{LLM\_request('question', 'response format')}$ directly within the generated code. The LLM should generate the "question" and "response format" according to the function to be implemented. The predefined $\texttt{LLM\_request}$ function sends the message to the LLM and returns its response, enabling the code to dynamically leverage the LLM's knowledge.

Additionally, the feedback and suggestions returned by each code-based rule are automatically generated by prompting the LLM with the corresponding rule. The detailed prompts used to generate these code-based rules can be found in Appendix \ref{sec:Translate Natural Language Rules to Code}. These feedback and suggestions play a crucial role in helping the agent refine and improve its planning process (Section \ref{sec:Inference on LLM Agents with Learned Rules}).

\begin{center}
    \textbf{Code-based Rules for Minecraft}
\end{center}

\lstset{language=}
\begin{lstlisting}
def Rule_3_craft(state, action):
    if action['name'] == 'craft':
        obj = list(action['args']['obj'].keys())[0]
        platform = action['args']['platform']
        
        # Ask the LLM if the specified platform is required for the object being crafted
        question = f"Is a specific platform required to craft {obj} in Minecraft? If yes, what is the platform?"
        response_format = "only reply with the platform name (e.g., 'crafting table', 'furnace') or 'None' if no specific platform is required"
        required_platform = LLM_request(question + response_format)
        
        if required_platform != 'None' and platform != required_platform.lower():
            feedback = f"Crafting {obj} requires a {required_platform}, but {platform} was provided."
            success = False
            suggestion = f"Use a {required_platform} to craft {obj}."
            return feedback, success, suggestion
        else:
            feedback = f"Crafting {obj} was successful."
            success = True
            suggestion = ""
            return feedback, success, suggestion
    else:
        feedback = "Action type is not 'craft', so it is considered successful."
        success = True
        suggestion = ""
        return feedback, success, suggestion

def Rule_6_craft(state, action):
    feedback = ""
    success = True
    suggestion = ""

    if action["name"] == "craft":
        materials_needed = action["args"]["materials"]
        inventory = state["inventory"]

        for material, quantity in materials_needed.items():
            if inventory.get(material, 0) < quantity:
                feedback = f"Failed to craft {list(action['args']['obj'].keys())[0]} due to insufficient {material}."
                success = False
                suggestion = f"Gather more {material} to craft {list(action['args']['obj'].keys())[0]}."
                break
        else:
            feedback = f"Successfully crafted {list(action['args']['obj'].keys())[0]}."

    return feedback, success, suggestion

def Rule_2_mine(state, action):
    feedback = ""
    success = True
    suggestion = ""

    if action["name"] == "mine":
        obj = list(action["args"]["obj"].keys())[0]
        tool = action["args"]["tool"]

        # Check if the tool is appropriate for the object being mined
        question = f"Is the tool '{tool}' appropriate for mining '{obj}' in Minecraft? Only reply True or False."
        is_tool_appropriate = LLM_request(question)

        if is_tool_appropriate == "False":
            feedback = f"The tool '{tool}' is not appropriate for mining '{obj}'."
            print(feedback)
            success = False
            suggestion = f"Use an appropriate tool for mining '{obj}'."
        else:
            feedback = f"The tool '{tool}' is appropriate for mining '{obj}'."
            print(feedback)
            success = True

    return feedback, success, suggestion

def Rule_7_mine(state, action):
    feedback = ""
    success = True
    suggestion = ""

    if action["name"] == "mine":
        tool = action["args"]["tool"]
        if tool and tool not in state["inventory"]:
            feedback = f"Action failed: Tool '{tool}' is not in the inventory."
            success = False
            suggestion = f"Please ensure you have the '{tool}' in your inventory before mining."
        else:
            feedback = "Action succeeded: Tool is present in the inventory."
            success = True
            suggestion = ""

    return feedback, success, suggestion

def Rule_2_gather(state, action):
    feedback = ""
    success = True
    suggestion = ""
    if action["name"] == "gather":
        sky_light_level = state["location_stats"]["sky_light_level"][0]
        if sky_light_level < 1.0:
            feedback = "Action failed: sky light level is less than 1.0."
            success = False
            suggestion = "Wait until the sky light level is higher."
        else:
            feedback = "Action succeeded."
            success = True
            suggestion = ""
    else:
        feedback = "Action succeeded."
        success = True
        suggestion = ""
    return feedback, success, suggestion
    
def Rule_1_fight(state, action):
    # Extract action name and arguments
    action_name = action.get("name")
    action_args = action.get("args", {})
    
    # Initialize feedback, success, and suggestion
    feedback = ""
    success = True
    suggestion = ""
    
    # Rule 1: For action 'fight', check if the 'tool' is present in 'inventory' or 'equipment'
    if action_name == "fight":
        tool = action_args.get("tool")
        if tool:
            inventory = state.get("inventory", {})
            equipment = state.get("equipment", {})
            if tool not in inventory and tool not in equipment:
                feedback = f"Action '{action_name}' failed: Tool '{tool}' is not present in inventory or equipment."
                success = False
                suggestion = f"Ensure the tool '{tool}' is available in either inventory or equipment before attempting to fight."
            else:
                feedback = f"Action '{action_name}' succeeded: Tool '{tool}' is available."
        else:
            feedback = f"Action '{action_name}' failed: No tool specified."
            success = False
            suggestion = "Specify a tool to use for the fight action."
    else:
        feedback = f"Action '{action_name}' is not restricted by the rule."
    return feedback, success, suggestion
\end{lstlisting}

\begin{center}
    \textbf{Code-based Rules for ALFWorld}
\end{center}

\lstset{language=}
\begin{lstlisting}
def Rule_1_clean(state, action):
    if action['name'] == 'clean':
        obj_to_clean = action['args']['obj']
        item_in_hand = state['item_in_hand']['item_name']
        if obj_to_clean != item_in_hand:
            feedback = f"Action failed: {obj_to_clean} is not in hand."
            success = False
            suggestion = f"Please take {obj_to_clean} in hand before cleaning."
            return feedback, success, suggestion
    feedback = "Action executed successfully."
    success = True
    suggestion = ""
    return feedback, success, suggestion

def Rule_3_clean(state, action):
    if action["name"] == "clean":
        tool = action["args"]["tool"]
        if tool not in state["reachable_locations"]:
            feedback = f"Action failed: The tool '{tool}' is not reachable."
            success = False
            suggestion = f"Make sure the tool '{tool}' is in the list of reachable locations."
            return feedback, success, suggestion
        else:
            feedback = "Action succeeded: The tool is reachable."
            success = True
            suggestion = ""
            return feedback, success, suggestion
    else:
        feedback = "Action succeeded: The action type is not 'clean'."
        success = True
        suggestion = ""
        return feedback, success, suggestion

def Rule_5_clean(state, action):
    if action['name'] == 'clean':
        current_position = state['current_position']['location_name']
        tool_location = action['args']['tool']
        if current_position != tool_location:
            feedback = f"Action 'clean' failed: You are not at the tool location ({tool_location})."
            success = False
            suggestion = f"Move to the tool location ({tool_location}) before cleaning."
            return feedback, success, suggestion
    # If the action is not 'clean' or the rule conditions are met
    feedback = "Action executed successfully."
    success = True
    suggestion = ""
    return feedback, success, suggestion

def Rule_2_take(state, action):
    feedback = ""
    success = True
    suggestion = ""
    if action["name"] == "take":
        if state["item_in_hand"]["item_name"] is not None:
            feedback = "Action failed: Agent is already holding an item."
            success = False
            suggestion = "You may heat, put, cool the item in hand directly without removing the other items in target location/container."
        else:
            feedback = "Action succeeded: Agent is not holding any item."
            success = True
            suggestion = ""
    else:
        feedback = "Action succeeded: Action type is not 'take'."
        success = True
        suggestion = ""
    return feedback, success, suggestion

def Rule_4_take(state, action):
    if action['name'] == 'take':
        current_location = state['current_position']['location_name']
        source_location = action['args']['source']
        if current_location != source_location:
            feedback = "Action failed: Agent is not at the location of the item."
            success = False
            suggestion = f"Move to {source_location} before taking the item."
            return feedback, success, suggestion
    # If the action is not 'take', it is considered successful
    feedback = "Action executed successfully."
    success = True
    suggestion = ""
    return feedback, success, suggestion

def Rule_3_open(state, action):
    if action['name'] == 'open':
        target = action['args']['target']
        current_position = state['current_position']['location_name']
        
        if current_position != target:
            feedback = f"Action 'open' failed: You are not at the target location '{target}'."
            success = False
            suggestion = f"Move to '{target}' before trying to open it."
            return feedback, success, suggestion
        else:
            feedback = f"Action 'open' succeeded: You are at the target location '{target}'."
            success = True
            suggestion = ""
            return feedback, success, suggestion
    else:
        feedback = "Action succeeded: The action type is not 'open'."
        success = True
        suggestion = ""
        return feedback, success, suggestion

def Rule_2_put(state, action):
    if action['name'] == 'put':
        item_in_hand = state['item_in_hand']['item_name']
        item_to_put = action['args']['obj']
        if item_in_hand != item_to_put:
            feedback = f"Action failed: The item '{item_to_put}' is not in hand."
            success = False
            suggestion = f"Please ensure you have '{item_to_put}' in hand before attempting to put it."
            return feedback, success, suggestion
    # If the action is not 'put', it is considered successful
    feedback = "Action executed successfully."
    success = True
    suggestion = ""
    return feedback, success, suggestion
        
def Rule_1_use(state, action):
    if action['name'] == 'use':
        obj = action['args']['obj']
        current_location = state['current_position']['location_name']
        # Check if the object is in the current location
        if obj not in state['items_in_locations'].get(current_location, []):
            feedback = f"Action failed: {obj} is not at the current position {current_location}."
            success = False
            suggestion = f"Move to the location where {obj} is present or bring {obj} to the current location."
            return feedback, success, suggestion
    # If the action is not 'use', it is considered successful
    feedback = "Action executed successfully."
    success = True
    suggestion = ""
    return feedback, success, suggestion

def Rule_1_heat(state, action):
    feedback = ""
    success = True
    suggestion = ""
    if action["name"] == "heat":
        tool = action["args"]["tool"]
        current_position = state["current_position"]["location_name"]
        if current_position != tool:
            feedback = f"Action failed: The tool '{tool}' is not at the current position '{current_position}'."
            success = False
            suggestion = f"Move to the location of the tool '{tool}' before attempting to heat."
        else:
            feedback = "Action succeeded: The tool is at the current position."
            success = True
            suggestion = ""
    else:
        feedback = "Action succeeded: The action type is not 'heat'."
        success = True
        suggestion = ""
    return feedback, success, suggestion
    
def Rule_5_heat(state, action):
    if action["name"] == "heat":
        item_in_hand = state["item_in_hand"]["item_name"]
        item_to_heat = action["args"]["obj"]
        
        if item_in_hand != item_to_heat:
            feedback = f"Action failed: You are trying to heat {item_to_heat} but you are holding {item_in_hand}."
            success = False
            suggestion = f"Hold {item_to_heat} before trying to heat it."
            return feedback, success, suggestion
        
    feedback = "Action executed successfully."
    success = True
    suggestion = ""
    return feedback, success, suggestion

def Rule_1_go_to(state, action):
    if action['name'] == 'go_to':
        current_location = state['current_position']['location_name']
        target_location = action['args']['target']
        
        if current_location == target_location:
            feedback = f"Action failed: You are already at {target_location}."
            success = False
            suggestion = "Try moving to a different location."
            return feedback, success, suggestion
        
    # If the action is not 'go_to' or the target location is different from the current location
    feedback = "Action executed successfully."
    success = True
    suggestion = ""
    return feedback, success, suggestion
\end{lstlisting}

\section{Experiment Details}

\subsection{Minecraft}
\label{sec:Minecraft Experiment Details}

\textbf{Task Details.} 
We used the 'Tech Tree' series tasks in MineDojo. Minecraft presents a structured progression system involving various levels of tools and armor, each with unique properties and increasing difficulty to unlock. To advance through these levels, the agent must develop and apply systematic, compositional skills to navigate the technology tree. 
Tasks are structured into six technology tiers: \textit{wood}, \textit{stone}, \textit{iron}, \textit{gold}, \textit{diamond}, and \textit{redstone} with each level presenting a higher degree of difficulty. And each level contains a certain number of tasks, which is shown in Table \ref{tab:Techtree Task Details}. Additionally, we only do tasks in overworld, so tasks that require materials from Nether and End to complete are disregarded (e.g. redstone observer, redstone lamp, redstone comparator).

\begin{table}[t]
\caption{Techtree Task Details}
\vspace{-0.5em}
\label{tab:Techtree Task Details}
\begin{center}
\renewcommand\arraystretch{1.1}
\resizebox{\linewidth}{!}{

\begin{tabular}{l>{\centering\arraybackslash}p{15cm}}
\toprule
\textbf{Task Level} & \textbf{Tasks}                                                                                                                                                                              \\ \midrule 
Wooden              & wooden sword,                                                                            wooden pickaxe, wooden axe, wooden hoe, wooden shovel                                              \\
Stone               & stone sword,                                                                            stone pickaxe, stone axe, stone hoe, stone shovel                                                  \\
Iron                & iron sword, iron pickaxe, iron axe, iron hoe                                            iron shovel, iron boots, iron chestplate, iron helmet, iron leggings                               \\
Golden              & golden sword, golden pickaxe, golden axe, golden hoe                                    golden shovel, golden boots, golden chestplate, golden helmet, golden leggings                     \\
Diamond             & diamond sword, diamond pickaxe, diamond axe, diamond hoe                                diamond shovel, diamond boots, diamond chestplate, diamond helmet, diamond leggings                \\
Redstone            & redstone block, redstone clock, redstone compass, redstone dispenser, redstone dropper  redstone piston, redstone torch, redstone repeater,redstone detector rail, redstone activator rail \\ \bottomrule
\end{tabular}
    }
\end{center}
\end{table}

\textbf{Method Setup.} 
We utilize GPT-4o as the backend for our method.
To rigorously assess the agent's performance, we initialize it in the "open-ended" mode—the most challenging and interactive environment available, analogous to survival mode. In this setting, the agent starts with an empty inventory and randomized seeds for both the environment and its starting position, requiring it to strategize effectively. Unlike creative or adventure modes, the agent must contend with the dynamic generation of hostile mobs, introducing additional complexity and difficulty.
Starting without any resources, the agent is forced to actively mine materials and craft essential items to progress, testing its planning, adaptability, and problem-solving skills. 

We select four tasks from each task level to serve as the testing set and the remaining tasks to construct the training set. For the level with a limited number of tasks, such as \textit{Wood} and \textit{Stone}, we add additional tasks from Optimus-1~\citep{li2024optimus} to ensure sufficient diversity for rule learning in the training process. Finally, we have a total of 30 training tasks and 24 testing tasks.

Within MPC framework, reward is assigned as follows: a reward of $+1$ if the world model $f_{\text{wm}}$ predicts the transition will be successful ($\text{action\_result}=\text{True}$), and $0$ if it predicts failure ($\text{action\_result}=\text{False}$). The world model provides feedback to the agent, enabling the agent to refine its plan based on the state prior to the failed action and the received feedback. This iterative process continues until the task is successfully completed within the planning phase.

For buffered trajectories (e.g., GITM~\citep{zhu2023ghost}), we adopted the original settings by storing successful task trajectories. During planning, we search this buffer for the trajectory most similar to the current task and include it in the prompt as a reference.

\subsection{ALFWorld}
\label{sec:ALFWorld Experiment Details}

\textbf{Task Details.} ALFWorld is a virtual environment designed as a text-based simulation where agents perform tasks by interacting with a simulated household. The environment includes six distinct task types, each requiring the agent to accomplish a high-level objective, such as placing a cooled lettuce on a countertop. Agents use text commands to navigate and manipulate objects in the virtual space, for example, issuing instructions like "go to countertop 1," "take lettuce 1 from countertop 1," or "cool lettuce 1 with fridge 1." The visual observations from the agent’s point of view are converted into natural language descriptions before being delivered to the agent. The agent's state is represented by the cumulative history of these observations. Success is measured by the completion the specified task goal.

\textbf{Method Setup.} 
We conducted rule learning on the training set, with the resulting rules presented in Appendix \ref{sec:Learned Rules}. 
Since tasks in ALFWorld require agents to continuously gather information from the environment, and our learned rules focus on capturing the dynamic of the environment, we adopted a one-step MPC. This method evaluates whether the agent's current action aligns with the environment's dynamic patterns based on its state information.
Additionally, to enhance rule discovery, we developed scripts to convert the natural language dialogue history and action information into a structured JSON format, as illustrated in Appendix \ref{sec: ALFworld Environments' State Space and Action Space}. 
We utilize GPT-3.5-Instruct as our backbone model.

\subsection{Experiment Design for Effectiveness of Rule Learning}
\label{sec:Experiment Design for Effectiveness of Rule Learning}

We conduct 3 training tasks per iteration over a total of 10 iterations during training. After each iteration, the model, equipped with latest learned rules or buffered trajectories, is tested on the testing set.

The cover rate quantifies the extent to which the rules derived from the rule learning process address the LLM's failed predictions. 
Specifically, it represents the probability that mispredicted transitions by the LLM are correctly handled by the learned rules.

To assess the alignment between the LLM-based world model and the actual environment, 
we first identify transitions where the LLM fails to make accurate predictions. This is achieved by utilizing an unaligned LLM world model—one without any rules—to generate predictions for trajectories obtained from the test set. The discrepancies between the predicted states \( \hat{s}_{t+1} \) and the actual states \( s_{t+1} \) are compiled into a dataset of mispredicted transitions.
These mispredictions highlight areas where the LLM world model does not align with the environment's dynamics.

Subsequently, the learned rules at each iteration are evaluated against the mispredicted transitions dataset to determine their effectiveness in correcting these mispredictions.
If a rule successfully predicts the outcome of a previously mispredicted transition, it demonstrates that the rule effectively addresses the LLM's failure in that instance. The cover rate is then calculated as the ratio of correctly addressed mispredictions to the total number of mispredicted transitions:
\begin{equation}
    \text{Cover Rate} = \frac{\text{Number of Mispredictions Addressed by Rules}}{\text{Total Number of Mispredicted Transitions}}
\end{equation}
Furthermore, as depicted in Figure \ref{fig:rulelearningprocess} , predictions and rules are categorized by action types—\textit{craft}, \textit{mine}, \textit{gather}, and \textit{fight}—allowing the cover rate to be calculated for each action category individually.
A higher cover rate indicates that the rule learning process effectively enhances the alignment of the LLM world model with the environment, thereby improving the overall accuracy and reliability of the agent's planning.

\section{Greedy Algorithm}
\label{sec:Greedy Algorithm}

We implement the following Algorithm \ref{alg:Greedy Algorithm} to solve the maximum set cover problem \ref{eqn:maximum set cover problem}.

\begin{algorithm}[H]
\caption{Greedy Algorithm for Maximum Set Cover Problem}
\label{alg:Greedy Algorithm}
\begin{algorithmic}[1]
\State \textbf{Input:} 
\State $\mathcal{D}^{\text{incorrect}} = \{ \delta^{\text{incorrect}}_1, \delta^{\text{incorrect}}_2, \ldots, \delta^{\text{incorrect}}_n \}$ \Comment{Set of incorrect transitions to cover}
\State $R^{\text{code}} = \{ R^{\text{code}}_1, R^{\text{code}}_2, \ldots, R^{\text{code}}_m \}$ \Comment{Set of rules covering subsets of transitions}
\State $a_{ij}$: Indicator matrix where $a_{ij} = 1$ if $\delta^{\text{incorrect}}_j \in R^{\text{code}}_i$, otherwise $a_{ij} = 0$

\State \textbf{Output:} Set of selected rules $R_{\text{selected}}$

\State Initialize $R_{\text{selected}} \gets \emptyset$
\State Initialize $\mathcal{D}_{\text{covered}} \gets \emptyset$ \Comment{Set of covered transitions}
\State Initialize $x_i \gets 0$ for all $i \in \{1, \ldots, m\}$ \Comment{Rule selection indicators}
\State Initialize $y_j \gets 0$ for all $j \in \{1, \ldots, n\}$ \Comment{Transition coverage indicators}

\While{$\mathcal{D}_{\text{covered}} \neq \mathcal{D}^{\text{incorrect}}$}
    \State For each rule $R_i^{\text{code}} \in R^{\text{code}}$, compute:
    \[
    \text{gain}(R_i) = \left| \left( \{ \delta^{\text{incorrect}}_j \mid a_{ij} = 1 \} \setminus \mathcal{D}_{\text{covered}} \right) \right|
    \]
    \State Select the rule $R_i^{\text{code}}$ with the largest gain, i.e.,
    \[
    i^* = \arg\max_i \text{gain}(R_i)
    \]
    
    \If{$\max \text{gain}(R_i) = 0$}
        \State \textbf{Break} \Comment{Terminate if no rule can cover any additional transitions}
    \EndIf

    \State Add $R_{i^*}^{\text{code}}$ to $R_{\text{selected}}$
    \State Update $\mathcal{D}_{\text{covered}} \gets \mathcal{D}_{\text{covered}} \cup \{ \delta^{\text{incorrect}}_j \mid a_{i^*j} = 1 \}$
    \State Set $x_{i^*} \gets 1$ \Comment{Mark rule $R_{i^*}^{\text{code}}$ as selected}
    \State For each $\delta^{\text{incorrect}}_j$ covered by $R_{i^*}^{\text{code}}$, set $y_j \gets 1$
\EndWhile

\State \textbf{Return} $R_{\text{selected}}$
\end{algorithmic}
\end{algorithm}

\section{Code-based rules verification logic}
\label{sec:verification logic}

\textbf{Criteria for Determining Whether a Rule is Active.}
We apply rules only in scenarios where they are relevant to the current transition, defining this relevance as the rule being "active" or "effective." 
A rule is considered active if its outcome of the current transition yields the specific outcome it is designed to address. Specifically, in our framework, transition success is represented as \texttt{True}, and failure is represented as \texttt{False}.
\begin{itemize}
    \item Rules Designed to Identify Successes:\\
    A rule intended to detect successes is considered active when it evaluates the current transition and returns \texttt{True}.
    \item Rules Designed to Identify Failures:\\
    A rule intended to detect failures is considered active when it evaluates the current transition and returns \texttt{False}.
\end{itemize}
In essence, a rule is active when its outcome aligns with the type of outcome it is meant to assess (either success or failure). This ensures that rules are applied appropriately and only influence the LLM world model's predictions when relevant to the specific circumstances of the transition.

\textbf{Determining Whether a Rule is Correct or Incorrect}
When a rule is active, if it makes an incorrect judgment—predicting success when the transition actually fails or vice versa—the rule is considered invalid and is removed from the rule set. Transitions where the rule is not applicable—referred to as "inactive" or "dormant"—are excluded from the evaluation process.

\section{Limitation and Future Work}

Currently, our rule learning framework generates simple rules that primarily assess whether actions align with environment dynamics (i.e., rules for transitions). Future research should explore advanced reasoning methods that enable LLMs to derive more abstract rules, such as those governing entire planning processes.
Furthermore, many embodied environments exhibit stochastic dynamics, where actions have probabilistic outcomes. For example, resource gathering at night in Minecraft often fails due to hostile creatures but can sometimes succeed. Our current rule learning process cannot handle such randomness, typically classifying these scenarios as failures. Addressing this limitation by enabling rules to account for stochastic dynamics is a promising research direction, potentially leading to more accurate and reliable world models.


\end{document}